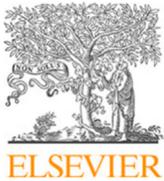
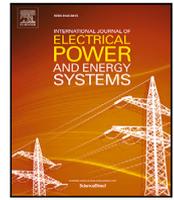

# Early wind turbine alarm prediction based on machine learning—Alarm Forecasting

Syed Shazaib Shah  *, Tan Daoliang

*School of Energy and Power, Beihang University, Beijing, 100191, PR China*

ABSTRACT

Alarm data is pivotal in curbing fault behavior in Wind Turbines (WTs) and forms the backbone for advanced predictive monitoring systems. Traditionally, research cohorts have been confined to utilizing alarm data solely as a diagnostic tool—merely indicative of unhealthy status. However, this study aims to offer a transformative leap towards preempting alarms, preventing alarms from triggering altogether, and consequently averting impending failures. Our proposed Alarm Forecasting and Classification (AFC) framework is designed on two successive modules: first, the regression module based on long short-term memory (LSTM) for time-series alarm forecasting, and thereafter, the classification module to implement alarm tagging on the forecasted alarm. This way, the entire alarm taxonomy can be forecasted reliably rather than a few specific alarms. 14 Senvion MM82 turbines with an operational period of 5 years are used as a case study; the results demonstrated 82%, 52%, and 41% accurate forecasts for 10, 20, and 30 min alarm forecasts, respectively. The results substantiate anticipating and averting alarms, which is significant in curbing alarm frequency and enhancing operational efficiency through proactive intervention.

## 1. Introduction

The renewable energy sector has become increasingly vital in addressing global climate challenges due to international commitments such as the Paris Agreement [1]. This has resulted in a global transition from finite fossil fuels to sustainable alternatives like wind and solar energy [2,3]. The wind energy sector has experienced remarkable growth over the past two decades [4], driven by three key developments: (1) advanced control systems [5], (2) more resilient turbine designs [6], and (3) steadily declining costs [7]. According to International Renewable Energy Agency (IRENA) [8] projections, these advancements will propel global installed capacity to 1000 GW by 2050. The widespread adoption of Supervisory Control and Data Acquisition (SCADA) systems has been transformative for wind energy, enabling real-time monitoring of turbine components through comprehensive sensor networks [9–11]. These systems acquire 50+ operational parameters including temperature, power output, and wind characteristics [12–15], while triggering alarms for threshold violations [16]. Modern SCADA capabilities now support: (1) Predictive maintenance through fault pattern recognition [17,18]. (2) Anomaly detection via machine learning [19,20]. (3) Automated control optimization [21]. Recent advances integrate SCADA with digital twins for enhanced failure prediction [22], addressing limitations in traditional threshold-based alarms.

On note of SCADA use, a significant research cohort has been dedicated to leveraging alarm data for better control operations, increasing reliability, and enabling early fault detection and anomaly prediction [19,21]. However, although alarm-based methodologies are prevalent, they remain confined to the diagnostic realm, meaning based on the alarm readings, a classification task is employed to evaluate whether the turbine is in a healthy or unhealthy state [23–26]. While this approach can be effective in identifying faulty behavior, it is challenging to prevent such faults from occurring. This limitation stems from their reliance on real-time alarm data, as alarms are only triggered when the fault progression has been initiated. However, if these same alarms can be known in advance, it will allow for a sufficient time window to initiate a countermeasure response by regulating the relevant parameters via the control system, thereby preventing these impending alarms. This is the motivation behind this study: Alarm Forecasting for Wind Turbines (WTs), specifically in a range of 10–30 min windows. This proactive approach also helps reduce false alarms — a major challenge for turbine reliability — by allowing early detection and even correction of incoming alarms. As a result, false alarms can be identified and made less frequent.

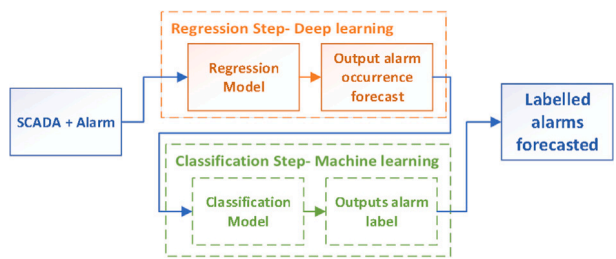

**Fig. 1.** Bird's-eye-view of AFC methodology.

Currently, SCADA/alarm-based approaches can be broadly divided into two main categories: Regression-based, employed to achieve time-series predictions [27–29]; and Classification-based, engineered to distinguish between the healthy and unhealthy status of the turbine [30–32]. However, alarm forecasting poses unique challenges; contrary to fault predictions, which have been limited to using real-time alarm data, the transition from real-time alarm data to forecasted one is inherently a different feat. For one reason, the sheer abundance of alarms makes it a daunting task. To achieve effective forecasting, the alarms need to be accurately predicted in a given future window, while at the same time, the specific alarm codes have to be reliably ensured. This dual-task methodology — combining alarm forecasting and classification — transcends conventional fault diagnosis approaches that merely distinguish between faulty and normal states. To achieve this one needs to simultaneously achieve:

1. Temporal forecasting of alarm occurrences (regression models supersedes such domain [33,34]).
2. Precise identification of specific alarm types (classification supersedes this domain [35,36]).

Therefore, based on this analogy, our study employed a hybrid approach coupling a recurrent neural network (RNN) (a multi-layered long short-term memory (LSTM) model) with statistical Machine Learning (ML) models; named AFC. Given that the SCADA alarms are short-term indicators [19], this study envisioned forecasting incoming alarms within a 10 to 30 min time window, having significance in providing ample time for an automated control system response and avoiding their occurrence. Short-term horizons of 10–30 min align with SCADA data resolution, operational decision-making cycles, and grid-balancing requirements. Such windows leverage existing data granularity (10 min averages), meet the latency constraints of turbine control systems, and provide actionable lead time for maintenance and grid dispatch adjustments [37]. The control system-initiated response is emphasized due to the short-lived nature of alarms and the accessibility challenges associated with them.

The contributions of this study are enumerated as follows:

1. A preemptive methodology is presented that leverages the strengths of both regression and classification ML techniques, combining time-series alarm forecast through LSTM regression with alarm nomenclature identification using ML classifiers.
2. Incoming alarms are forecasted 10–30 min in advance, paving the way for preemptive action.
3. Our AFC approach not only excels in conventional alarm detection tasks but also supersede by a high margin to all other state-of-the-art models in forecasting alarms.
4. Moreover, it is important to point out here that the dataset acquired for this study was in a very poor state; therefore, a series of meticulous data-preprocessing steps were taken to make it usable. These steps could be valuable tools for future researchers to make use of WT SCADA datasets since many of the SCADA datasets are often in similar poor condition.
5. By forecasting alarms in advance and rectifying their underlying cause before their manifestation, the potential to curtail the excessive false alarm issue is established.

The remainder of the paper follows: Section 2 gives a framework for the AFC methodology; discussion on the used dataset and its meticulous pre-processing steps for forecasting tasks are delineated in this section. This section also provides some key innovative ways to address the notorious poor quality dataset issues (excessive NaN values); Section 4 is dedicated to results and discussion; and finally, Section 5 concludes the AFC contributions. Section 6 points towards future work.

## 2. AFC methodology framework

The concept for AFC is rooted in a straightforward principle: leveraging the strengths of regression and classification-specific ML models to enhance time-series forecasting and classification for WT alarms. First, a LSTM model is employed, chosen through empirical testing, to forecast whether an alarm will occur within the next 10–30 min. The LSTM model will generate binary markers, '0' for no alarm and '1' for an alarm. Once impending alarms (binary markers) are forecasted, these are passed to a bagged-classifier consisting of three high-fidelity classification models of k-nearest neighbors (KNN), decision tree (DT), and random forest (RF), which classifies the binary indicators into their respective alarm codes, thus achieving finalized tagged incoming alarms. A bird's-eye-view of our adopted methodology, AFC, is depicted in Fig. 1. The rationale behind having this dual-task methodology in a series framework is that it is quite challenging, or impossible to achieve forecasting of alarms with a single standalone model. This is substantiate as of following reasons: (1) The shear number of alarms in a given turbine can overwhelm any model; the exact number of a given turbine system can vary but for reference, [38] mentions 368 defined unique alarms and the occurrence frequency can reach up to several thousands in just a span of 10 min, making it challenging to achieve forecasting. (2) Most standalone models can either handle regression, which is needed for time-series forecasting, or they can be optimized for classification, like finding the exact alarm—alarm tagging. However, as mentioned, due to the sheer number of alarm frequencies and their very close correlation among different alarms, a single model can be fooled. Therefore, to make alarm forecasting conducive, through empirical testing, this dual-task regression-classification series framework of AFC was envisioned. To substantiate the arguments made here, we also presented the validation results (Section 4.6) from some of the recently used methodologies, where most models were able to achieve respectable results when simply doing the classification task of alarm tagging; however, as we added the forecasting task on top of it, the models' performance plummeted.

LSTM-based regression models are particularly well-suited and have high fidelity for time-series forecasting tasks due prowess in effectively capturing long-term dependencies and complex temporal patterns in sequential data [39,40]. On the other hand, bagging techniques incorporating the statistical classifier models as the ones used for this study, like KNN, DT, and RF, have been in the limelight for a cost-effective approach towards WT alarm classification [41–43]. Bagging (bootstrap aggregating) significantly reduces variance, improves stability, and enhances predictive accuracy by averaging multiple base learners trained on resampled subsets of the data [44]. This approach is particularly effective for unstable classifiers (e.g., DTs) and can even bolster performance in high-variance models like KNN, yielding robustness to noise and outliers and better generalization on unseen data.

For attestation of AFC, 14 Senvion WTs SCADA data (details in Section 2.1) was used to run the tests. The data contained relatively high impurities, as most real-world wind farm data do. The dataset was meticulously refined with an innovative approach, which is mentioned in Section 2.2. Afterwards, the data is preprocessed for forecasting application in Section 2.3. Thereafter, the model frameworks for the regression and the classifiers are briefed in Section 2.4.

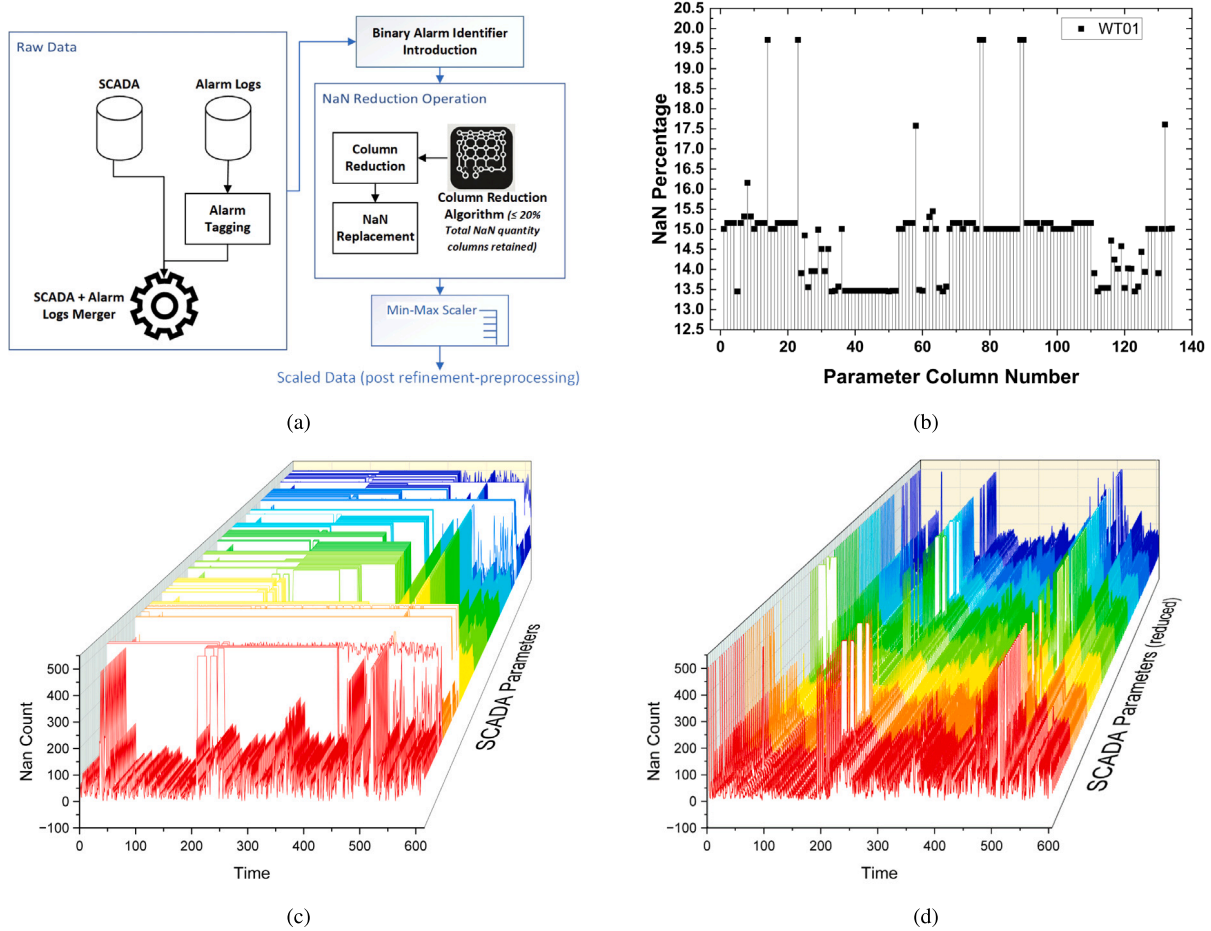

**Fig. 2.** (a) Overview of data preprocessing: Refinement. (b) Percentage of NaN for each parameter is shown for WT01. (c) Overview of time-series NaN values: WT01. The *x*-axis displays a scaled time parameter of 500 log entries, or, in other words, 83.3 h of accumulated time. The *y*-axis represents the number of log entries over this window, while the *z*-axis displays the sequential SCADA parameters order. The 3D graphical illustration gives an overview of how much NaN quantity of data was recorded. As seen in the graph, for some parameters, the plot is reaching the limit of 500 on the *y*-axis, which means for those parameters or periods of operation, either the WT was in an irregular state or the sensor associated with that parameter was faulty. (d) Post-NaN-reduction operation. Time-series view of the NaN values of the retained 136 parameters with ≤20% NaN values for WT01 dataset.

## 2.1. Data procurement and analysis

Acquiring a suitable SCADA dataset could be a daunting task due to accessibility and confidentiality concerns in the wind industry [45, 46]. Companies guard their SCADA data to prevent competitors from gaining insights. The inclusion of alarm logs can further exacerbate the confidentiality concern by highlighting intricate shortcomings. Combining both SCADA with the alarm logs is rare. The AFC methodology uses a dataset from the Penmanshiel wind farm in the UK, including static data on turbine coordinates with SCADA data and event logs from the alarm system from 14 Senvion MM82 turbines tagged as WT01 to WT15, with the exclusion of turbine 3 tagged WT03 due to its absence. The total operational length of the turbines was from 2016-06-06 to 2021-07-01, a total of approximately 5 years. The data was collected from a secondary system, Greenbyte, and some signals were unavailable due to sensor-faulty operation or irregularities. Interested readers can access this dataset through the provided link: https://github.com/sltzgs/Wind_Turbine_SCADA_open_data. The dataset has been made available by Cubico Sustainable Investments Ltd (https://www.cubicoinvest.com) under a CC-BY-4.0 open data license (https://creativecommons.org/licenses/by/4.0/legalcode) and is supplied in its current state.

Alarms logs were provided in a separate log file, consisting information such as the duration for a given alarm, alarm initiation time, alarm code, alarm description, alarm category, International Electrotechnical Commission (IEC) standard-based alarm classification, and auto-generated messages and comments. A total of 223 unique alarm codes were identified throughout the entire span of datasets.

To understand the given quality of Penmanshiel dataset, a thorough data analysis is done, which revealed a substantial number of NaN values. A 3D graph in Fig. 2(c) shows the incidence of NaN recorded values over time, showing significant gaps and potential issues during the WT operation. Of course, the data set in its current state is impractical for any application. This is further exacerbated by differences in NaN values across 14 turbine datasets. As each turbine is equipped with its own distinct sensors for collecting SCADA data, if one sensor is defective, of course, it does not imply that the same sensors for other turbines are also faulty. However, this does lead to considerable variance in the NaN values distribution across the chronological timelines for all WT datasets. This is critical because both RNNs and other ML models typically require datasets with consistent parameter types and sequences to ensure reliable training and prediction outcomes [47,48], posing a significant challenge in addressing and mitigating the impact of these NaN values without compromising its integrity for ML training. For instance, if a sensor in a particular WT is defective, the associated SCADA parameters linked to that sensor can be removed; however, this would necessitate the removal of the same healthy sensor data from other turbine's datasets to maintain uniformity across all datasets. Failure to do so would result in inconsistencies in the sets of SCADA parameters, which could render the datasets ineffective for training RNNs. To resolve the issue, NaN values must be eliminated while maintaining consistency among SCADA parameters across all datasets; thus, there is a need for specialized NaN-reduction steps. The next section delineates necessary sequential pre-processing steps, which are taken to make the dataset contingent with the AFC application, including the NaN-reduction steps mentioned earlier.

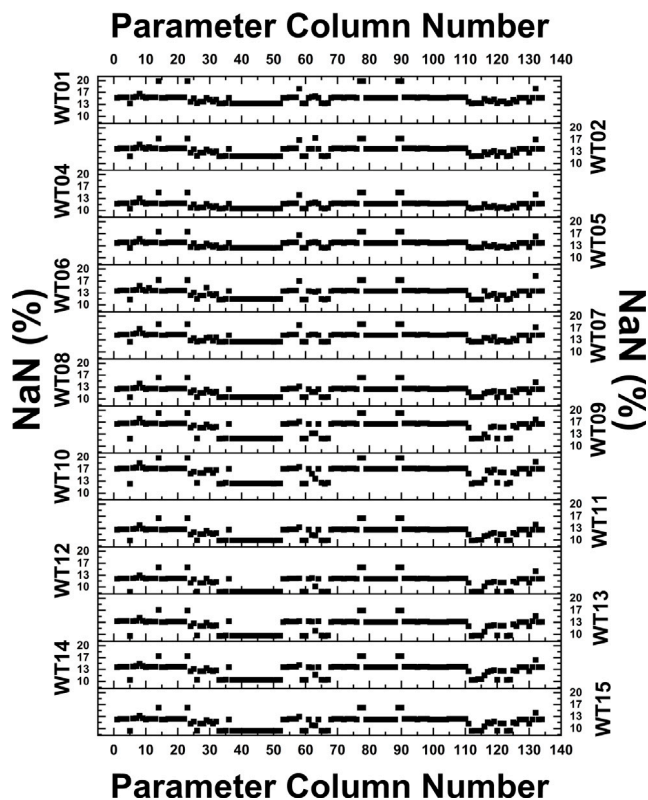

**Fig. 3.** All datasets parameter-wise NaN percentage: Each parameter is represented by its column number in the reduced form of the dataset along the *x*-axis.

*2.2. Data preprocessing: Refinement*

Besides the confidentiality concerns, WT SCADA tends to have sporadic NaN values, removal of which can be quite challenging, with the propensity to make the data set suboptimal. Researchers tend to operate various data cleansing techniques; conventionally, the effects tend to be mitigated by averaging over a larger temporal window [49,50]. This could be acceptable if the objective pertains to conditional monitoring, as the amalgamate NaN effect can be proportionally inconsequential. However, this is not the case for AFC. Our methodology relies on a short-term contingency of time-series progression; each preceding one to three logs contributes towards determining whether a given time-step yields an alarm or not. For such a short-term window, averaging techniques are non-pertinent. Perhaps this has limited the use of SCADA to short-term forecasting applications by other researchers. Therefore, by implementing meticulous preprocessing steps, our study refined datasets for AFC application and addressed the challenges of short-term applications of SCADA. Fig. 2(a) illustrates the sequential workflow required to preprocess and format the raw SCADA data for ML model training. These preprocessing steps are essential for refining the datasets, ensuring they are prepared for effective use in the subsequent stages of model development.

*2.2.1. Alarm tagging*

Prior to the merger operation of alarm and SCADA, each alarm entry was numerically labeled/tagged for ML model training. The alarm raw data did include designated numeric alarm codes for each alarm. However, these codes lacked structured ordering, leading to highly skewed scales, which could be problematic from ML model training [51,52]. To mitigate this bias and ensure equitable treatment of alarm data, a re-tagging operation was done, assigning ascending numeric values ranging from 1 to 223. It is to point out here that the alarm log entries with the tag '0' in the raw data were omitted from the alarm logs, as they were associated with the normal operation of WT and were not included in the 223 unique alarm logs.

*2.2.2. Time-series data preparation*

The initial phase involved integrating alarm logs with SCADA data to analyze temporally associated events. With 14 datasets tagged from WT01 to WT15, each dataset can be expressed using the following equation.

$$\delta_i = \{\delta_1, \delta_2, \ldots, \delta_N\} \quad (1)$$

The *N* represents the total data logs for a specific WT dataset, typically around 300,000 log entries, and *i* denotes the specific failure dataset, ranging from 1 to 15, excluding 3.

Each data point $\delta_u$ from Eq. (1) represents a single row and each row contains $\rho$ parameters:

$$\delta_u = \{p_{u,1}, p_{u,2}, \ldots, p_{u,\rho}\} \quad (2)$$

The value of $\rho$ is 300 and the variable *u* varies from 1 to *N*.

*2.2.3. Binary alarm identifier*

Since the fundamental idea behind AFC method is to predict alarm occurrence first — that is, whether an alarm is likely to occur or not in the future — and then categorize/tag that alarm using a classifier, an additional binary alarm identification is added. If for a given log in the time-series data, there is an alarm present, the binary alarm identifier will record a value of unity (1), otherwise zero (0) for no alarm presence, i.e., normal operation. This additional time-series binary alarm identifier would be set as the output for the regression-based model, responsible for the forecasting application.

*2.2.4. NaN based dataset reduction*

As mentioned previously, one of the challenging steps for the utilization of SCADA data is the presence of a huge quantity of NaN values. To mitigate this issue, a NaN-based reduction operation is applied; NaN values were removed while maintaining homogeneity among 14 turbines using a reference dataset, WT01. This dataset was chosen for its ability to retain the most number of parameters among all 14 datasets. An allowable threshold of 20% NaN values was selected, retaining all parameters with a NaN value quantity of ≤ 20% for the entire operational length; reduced form shown in Fig. 2(d). NaN distribution post-reduction operation for remaining turbines is provided in Appendix. The initial 300 parameters were reduced to 136 after a NaN-based reduction operation, resulting in the value of $\rho$ in Eq. (2) to be 136, with the parameter-wise NaN quantity percentage for WT01 (reference dataset) is shown in Fig. 2(b).

Based on the remaining parameter tags from the reduced form of the WT01 dataset, the reduction operation for the remaining datasets is conducted by retaining the same 136 parameters for the other datasets. Fig. 3 shows the cumulative percentage for NaN values for all parameters in each WT dataset; all parameters fall within the 20% acceptable limit for NaN quantity. By following these NaN-reduction procedures, parameters with an excessive quantity of missing values were eliminated, mitigating potential issues during the training of ML models. It is important to acknowledge, however, that the dataset still contains a significant amount of NaN occurrences, falling short of an ideal scenario. The time-series distribution of NaN values for the first 15 parameters, post-NaN-reduction, across all datasets is depicted in Figs. 18, 19, and 20. In practice, achieving a completely clean and ideal dataset in real-world SCADA data collection is rare, and the current state of the data, though suboptimal, is reflective of the real scenario faced in industrial applications. This highlights the necessity of utilizing such advanced preprocessing techniques capable of handling substandard data for the training stages to ensure robust model performance. By following these steps, future researchers can deploy more practical approaches to refine and utilize SCADA datasets that are contingent on the actual scenario.

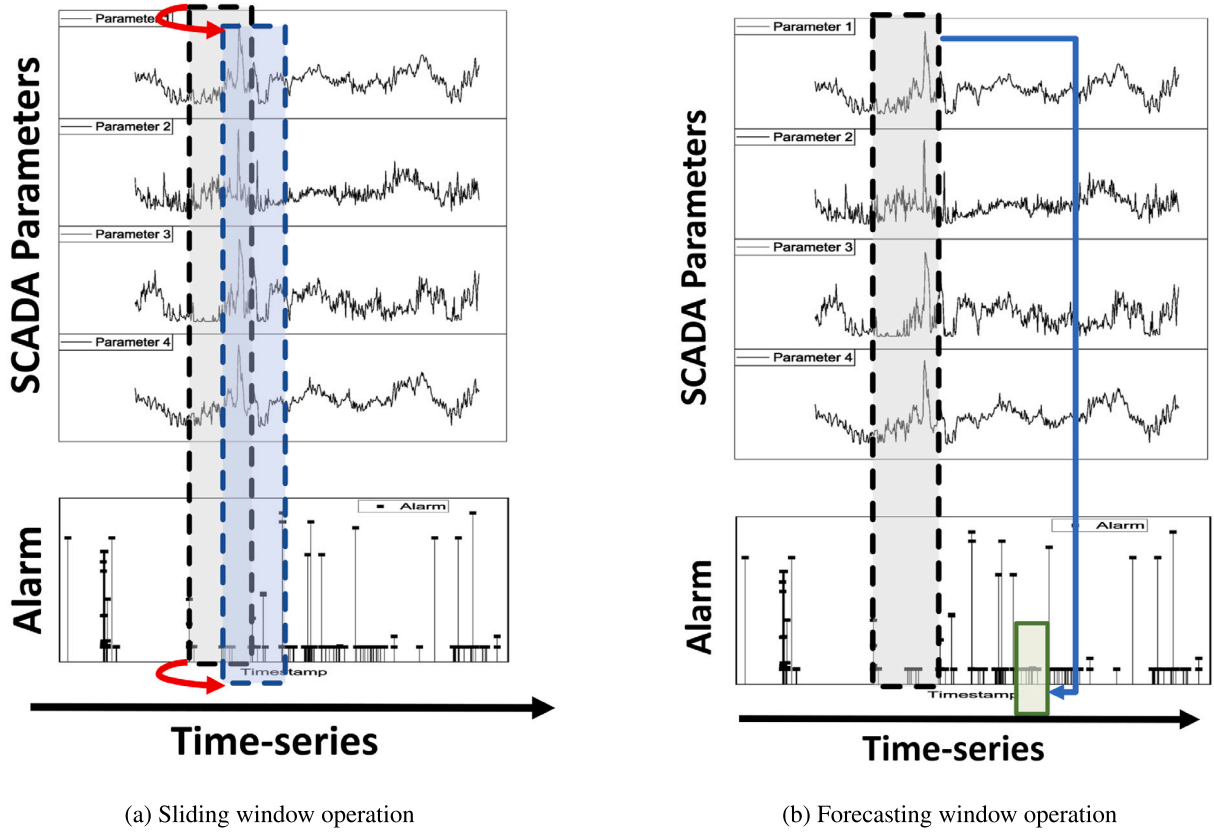

(a) Sliding window operation  (b) Forecasting window operation

**Fig. 4.** Comparison of sliding window and forecasting window operations.

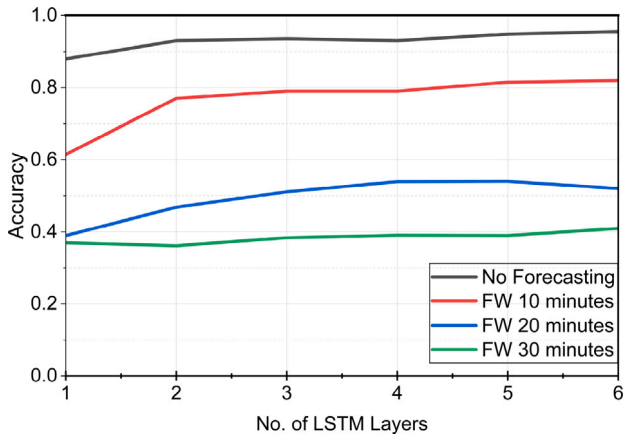

**Fig. 5.** LSTM layers appendant effect—Architecture deepening.

#### 2.2.5. Scaling

Each parameter $p_j$ is re-scaled, or better named as the normalization operation, using the minimum and maximum values specific to that parameter in order to get a scaled value $p_{u,j}^{\text{scaled}}$.

$$p_j = \{p_{1,j}, p_{2,j}, \ldots, p_{N,j}\} \quad (3)$$

$$p_{u,j}^{\text{scaled}} = \frac{p_{u,j} - \min(p_j)}{\max(p_j) - \min(p_j)} \quad (4)$$

where $p_{u,j}^{\text{scaled}}$ is the scaled value of $p_{u,j}$.

### 2.3. Data preprocessing: Alarm forecasting

After the removal of NaN values, the dataset undergoes additional preprocessing to reflect the time-series discrepancies and to incorporate modifications for forecasting; the sequential steps are briefed in the following subsections.

#### 2.3.1. Input and output data
1. **Input:** Since the AFC approach aims to discern discrepancies prompting alarm activation from the time-series SCADA data, the SCADA sensor readings plus the associated time-series alarm occurrence serve as the primary input for the LSTM-based forecasting step.
   **Input Data:**

$$\overline{X}_i = \{\overline{X}_{i,1}, \overline{X}_{i,2}, \ldots, \overline{X}_{i,N}\} \quad (5)$$

2. **Output 1: Binary Alarm Identifier**: The regression-based alarm forecasting system uses this metric as its output. The set contains a single column of binary form, with each '1' corresponding to an alarm instance. It is also important to point out here that this is the only output for which the forecasting technique would be applied; see Section 2.3.3.

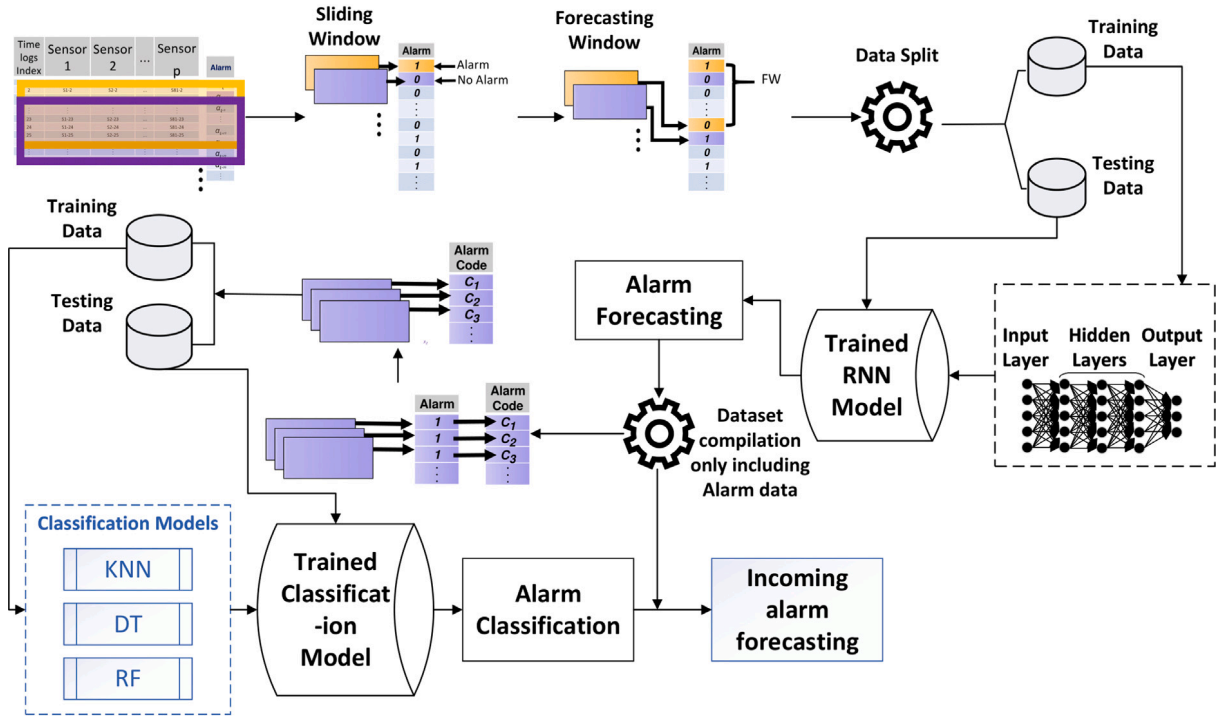

**Fig. 6.** AFC methodology schematic flowchart.

$$\overline{Y}_i = \{\overline{Y}_{i,1}, \overline{Y}_{i,2}, \ldots, \overline{Y}_{i,N}\} \quad (6)$$

3. **Output 2: Alarm Code**: This metric is set as the output for the classification-based part of this approach. It consist of the time-series logs of the alarm occurrence with each alarm code.

$$\overline{\overline{Y}}_i = \{\overline{\overline{Y}}_{i,1}, \overline{\overline{Y}}_{i,2}, \ldots, \overline{\overline{Y}}_{i,N}\} \quad (7)$$

*2.3.2. Sliding window application : 2D data structuring*

Following the input–output preprocessing step, the sliding window (SW) approach is employed to organize data for capturing long-term disparities. This technique involves structuring the dataset with a 2D window of fixed size sliding through the dataset and stacking adjacent data rows to produce a series of 2D data windows—with each window stepping one row (stride of one) from the previous. Each 2D-input unit generated post-SW implementation would be associated with the corresponding output value. To further understand this concept, see Fig. 4(a). The width and length of the SW are dependent on the number of rows (M = 136 for parameters and L = 12 for time logs), selected after testing, accuracy evaluation, and equipment capabilities constraints. The mathematical representation of the size of the SW ($SW_{Size}$) is denoted by Eq. (8).

$$SW_{Size} = SW_{Length} \times SW_{Width} \quad (8)$$

Following SW implementation, input and output datasets are structured as:

Input data for dataset *i*:

$$X_{input, i} = \{X_1, X_2, \ldots, X_{N-l}\}_i \quad (9)$$

Form the above equation, for a single input unit at an index $q$ i.e., $X_q$ where $q$ range from 1 to $N - l$, represents the 2D window of stacked rows of SCADA data—see the following equation:

$$X_{q,i} = \{d_u, d_{u+1}, \ldots, d_{u+l}\}_i \quad \text{where} \quad u \leq N - l \quad (10)$$

Output Data 1:

$$Y1_{output, i} = \{Y1_l, Y1_{l+1}, \ldots, Y1_N\}_i \quad (11)$$

for a given $Y1_w$ ($w$ ranging from $l$ to $N$) from the above equation, represents the output (binary identifier '1' or '0', signifying alarm occurrence) at index $w$. Also, as the AFC methodology includes a second ML step to achieve tagging operation, the output for that second step is to be equated as follows:

Output Data 2:

$$Y2_{output, i} = \{Y2_l, Y2_{l+1}, \ldots, Y2_N\}_i \quad (12)$$

To match up the input, output datasets lengths, few data logs would be omitted from the input and output, and for the sake of better understanding, the indices for the above equations would be revised. Eqs. (9), (11) and (12) are re-written as follows:

$$X'_{input,i} = \{X'_1, X'_2, \ldots, X'_R\}_i \quad (13)$$

$$Y1'_{output,i} = \{Y1'_1, Y1'_2, \ldots, Y1'_R\}_i \quad (14)$$

$$Y2'_{output,i} = \{Y2'_1, Y2'_2, \ldots, Y2'_R\}_i \quad (15)$$

where $R = N - l$. Since $l = 12$, it corresponds to a cumulative time period of 2 h (12 × 10 min). This means that for each input window, a time period of 2 h is used to determine whether there is an incoming alarm and, afterward, what kind of alarm it is going to be. Using the SW, an association between the last two hours of WT's behavior and the WT alarm is created. However, this association is still based on real-time prediction and will be predicting current alarms based on current turbine behavior; for future prediction, this association is to be manipulated to some steps ahead into the time series, elaborated in the next section.

*2.3.3. Forecasting window application: Forecasting introduction*

The SW application manipulates the association between input and output windows to predetermined steps ahead in time-series. The AFC application forecasts classified alarms 10–30 min in the future, with log-step sizes ranging from 1 to 3; named FW1, FW2 and FW3, respectively. Fig. 4(b) provides an illustration for the alarm forecasting task, illustrating the forecasting window (FW) application and defining the equations for post-FW induction. From Eqs. (13), (14) and (15), after implementing FW we get:

$$X_{f,i} = \{X'_1, X'_2, \ldots, X'_{R-f}\}_i \tag{16}$$

$$Y1_{f,i} = \{Y1'_f, Y1'_{f+1}, \ldots, Y1'_R\}_i \tag{17}$$

$$Y2_{f,i} = \{Y2'_f, Y2'_{f+1}, \ldots, Y2'_R\}_i \tag{18}$$

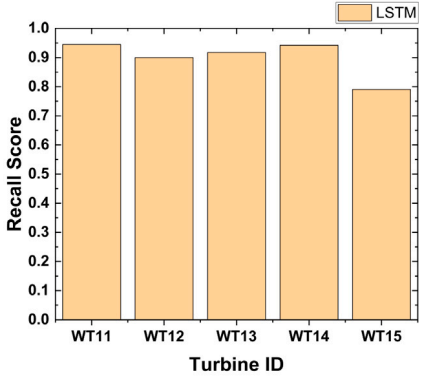

(a) FW: 10-minutes

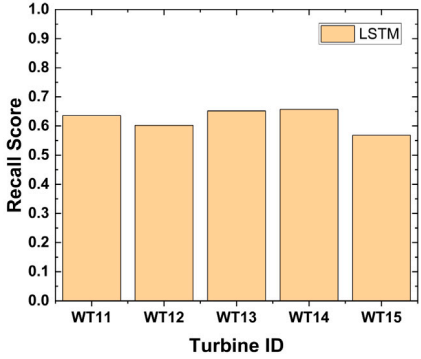

(b) FW: 20-minutes

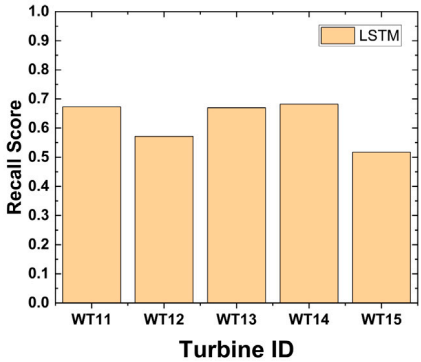

(c) FW: 30-minutes

**Fig. 7.** Recall Score—Regression model results: Regression-based binary alarm forecasting.

**Table 1**
Regression model architecture summary.

| Layer (type) | Output shape | Param # |
| --- | --- | --- |
| InputLayer (InputLayer) | (None, 12, 136) | 0 |
| LSTM (LSTM) | (None, 12, 512) | 1,329,152 |
| LSTM_1 (LSTM) | (None, 12, 256) | 787,456 |
| LSTM_2 (LSTM) | (None, 12, 128) | 197,120 |
| LSTM_3 (LSTM) | (None, 12, 64) | 49,408 |
| LSTM_4 (LSTM) | (None, 12, 32) | 12,416 |
| LSTM_5 (LSTM) | (None, 12, 16) | 3,136 |
| Reshape (Reshape) | (None, 12, 16) | 0 |
| Flatten (Flatten) | (None, 192) | 0 |
| Dense (Dense) | (None, 1) | 193 |

Total params: 2,378,881
Trainable params: 2,378,881
Non-trainable params: 0

The simplified concept from Eq. (16) to (18) is as follows:

$$\begin{bmatrix} X_{N-l} \to Y1_N \to Y2_N \\ X_{N-l-f} \to Y1_N \to Y2_N \\ (renaming\ for\ the\ sake\ of\ convenience,\ i.e., \\ X_{N-l-f} = X_{F,g}\ \&\ Y1_N = Y1_{F,g}\ \&\ Y2_N = Y2_{F,g}) \\ X_{F,g} \to Y1_{F,g} \to Y2_{F,g} \end{bmatrix}_i$$

It is quite important to specifically mention here that the FW effect if applied for $X_{N-l} \to Y1_N$ to make it into $X_{N-l-f} \to Y1_N$, and not for $Y1_N \to Y2_N$. After the FW application, the input data structure is complete. Now we need a custom deep neural network (DNN) model architecture that would be able to read through these forecasting-based temporal dependencies from the SCADA data to forecast alarm occurrences, which would eventually be classified into appropriate alarm tags using classifier models afterward. This is further detailed in the subsequent Section 2.4.

### 2.4. Model architecture

*Regression module*

The AFC application utilizes a RNN, specifically a LSTM network, designed for regression-based time-series binary forecast of alarm occurrence. As briefed in Section 2.3.3, the input for this model is the finalized form from the preprocessing step, i.e., Eq. (16); the model uses a multi-layer approach to gradually converge down to the desired outputs layer by layer; see Table 1. The model output, or prediction parameter, is set to Output 1 from Eq. (17), which aims to provide binary predictions for incoming alarms in a FW. Clusters of multiple alarms can be identified as unique alarm codes, but these approaches are more beneficial for accessing fault behavior in a WT; for our specific application, using the time-series occurrence of the alarm reading is sufficient. **Input Layer**: The input layer (`input_1`) accepts sequences of data with a shape of (`None, 12, 136`). This indicates that the model expects input sequences with 12 time steps and 136 features. **LSTM Layers**: The model consists of multiple LSTM layers stacked on top of each other. Each LSTM layer processes the input sequence and outputs a sequence of hidden states.

- `lstm`: This layer has 512 units.
- `lstm_1`: This layer has 256 units.
- `lstm_2`: This layer has 128 units.
- `lstm_3`: This layer has 64 units.
- `lstm_4`: This layer has 32 units.
- `lstm_5`: This layer has 16 units.

**Reshape Layer**: The `reshape` layer reshapes the output of the last LSTM layer to prepare it for the subsequent dense layer. It reshapes the output to (`None, 12, 16`). **Flatten Layer**: The `flatten` layer flattens the reshaped output into a one-dimensional vector (`None,` 192). **Dense Layer**: The `dense` layer is a fully connected layer with 1 unit, which produces the final output of the model (see Table 1).

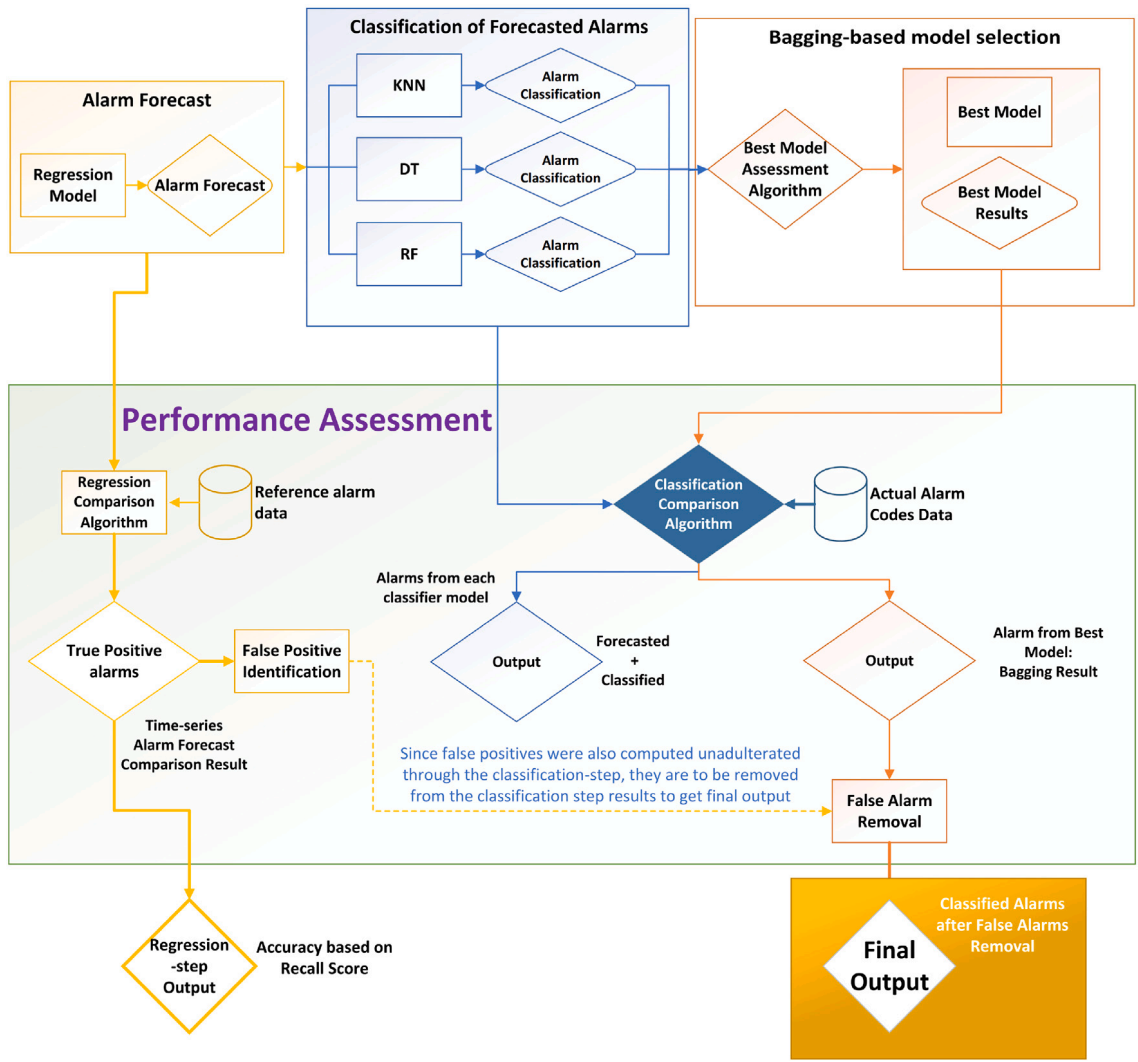

**Fig. 8.** Bagging-based model selection process.

The model comprises 2,378,881 parameters, including both trainable and non-trainable ones. Trainable parameters are updated during training to enhance performance, while non-trainable parameters remain fixed. A 2-hour worth of SCADA with time-stamped alarm logs is to be fed into the trained RNN model, which outputs whether there is an alarm occurrence in the next 10 to 30 min or not. The model layers were explicitly tuned through empirical testing. While for most applications, two or up to three layers usually suffice, however, during trial runs, we found the above configuration of 6 LSTM layers to give the best result. This inference was reached while accounting for the model performance across multiple scenarios up to the forecast window of 30 min; Fig. 5 shows the effect on accuracy for each appended layer. Around the 5th to 6th layer, the model shows negligible performance difference, plateauing the accuracy; therefore, the current 6-layer architecture was decided on based on this empirical evidence.

*Regression-classification transition*

Before we move towards the classification part of AFC, it is important to see through the transition process from the regression part to the classification part. The regression model will give us an output in binary form, i.e., '1' in case of an alarm and '0' in case of normal operation. Note that this is a time-series forecast of whether there is an incoming alarm or not. At this point, the code of the alarm is not known, i.e., the alarms are unclassified. Once these regression-based alarm forecasts are done, only the alarms '1' with their corresponding input window are taken. This window is then fed into the alarm-classifier to classify this alarm.

*Classification step for alarm classification task*

For this step, a multi-model ensemble approach was deployed; three well-known ML models were used, namely KNN, DT, RF (a bagged form of DT). The input will be fed to all these models in parallel, and whichever model gives the best result based on the recall score will be designated as the final output.

## 3. Training and testing

The model training and testing process is composed of two distinct stages: an initial regression phase followed by a classification phase; Fig. 6 provides an overview of the entire summary for AFC methodology. A standard testing ratio of 60–40 is used; WT01 to WT10, a total of 9 datasets, was used for the training applications, while WT11 to WT15, a total of 5 datasets, were used for the testing applications. The training and testing datasets maintain the time-series consistency and no random sampling was exercised. The successive steps taken for training-testing are as follow:

**Table 2**
Evaluation metrics for different forecast windows (FW1 and FW2) and turbines. The final score represents the post-FPAF removal.

| Forecast | Turbine | Evaluation parameter | LSTM | KNN | DT | RF |
|---|---|---|---|---|---|---|
| FW1 | WT11 | Precision | 0.7477 | 0.8141 | 0.9066 | 0.9370 |
| | | Recall | 0.9449 | 0.8141 | 0.9066 | 0.9370 |
| | | F1 Score | 0.8348 | 0.8141 | 0.9066 | 0.9370 |
| | | Final | 94.49% | 76.92% | 85.66% | 88.53% |
| | WT12 | Precision | 0.9537 | 0.7560 | 0.8083 | 0.8930 |
| | | Recall | 0.9000 | 0.7560 | 0.8083 | 0.8930 |
| | | F1 Score | 0.9260 | 0.7560 | 0.8083 | 0.8930 |
| | | Final | 90.00% | 68.04% | 72.74% | 80.36% |
| | WT13 | Precision | 0.8499 | 0.7992 | 0.8198 | 0.9084 |
| | | Recall | 0.9171 | 0.7992 | 0.8198 | 0.9084 |
| | | F1 Score | 0.8822 | 0.7992 | 0.8198 | 0.9084 |
| | | Final | 91.71% | 73.29% | 75.19% | 83.31% |
| | WT14 | Precision | 0.8069 | 0.8059 | 0.9289 | 0.9265 |
| | | Recall | 0.9423 | 0.8059 | 0.9289 | 0.9265 |
| | | F1 Score | 0.8693 | 0.8059 | 0.9289 | 0.9265 |
| | | Final | 94.23% | 75.94% | 87.53% | 87.31% |
| | WT15 | Precision | 0.9604 | 0.7273 | 0.6451 | 0.8702 |
| | | Recall | 0.7901 | 0.7273 | 0.6451 | 0.8702 |
| | | F1 Score | 0.8669 | 0.7273 | 0.6451 | 0.8702 |
| | | Final | 79.01% | 57.46% | 50.96% | 68.75% |
| FW2 | WT11 | Precision | 0.7262 | 0.6984 | 0.8780 | 0.8837 |
| | | Recall | 0.6359 | 0.6984 | 0.8780 | 0.8837 |
| | | F1 Score | 0.6781 | 0.6984 | 0.8780 | 0.8837 |
| | | Final | 63.59% | 44.41% | 55.83% | 56.19% |
| | WT12 | Precision | 0.8735 | 0.6405 | 0.7040 | 0.8272 |
| | | Recall | 0.6024 | 0.6405 | 0.7040 | 0.8272 |
| | | F1 Score | 0.7130 | 0.6405 | 0.7040 | 0.8272 |
| | | Final | 60.24% | 38.58% | 42.40% | 49.83% |
| | WT13 | Precision | 0.6708 | 0.6703 | 0.5657 | 0.8237 |
| | | Recall | 0.6520 | 0.6703 | 0.5657 | 0.8237 |
| | | F1 Score | 0.6613 | 0.6703 | 0.5657 | 0.8237 |
| | | Final | 65.20% | 43.70% | 36.89% | 53.71% |
| | WT14 | Precision | 0.6772 | 0.6791 | 0.8520 | 0.8585 |
| | | Recall | 0.6576 | 0.6791 | 0.8520 | 0.8585 |
| | | F1 Score | 0.6673 | 0.6791 | 0.8520 | 0.8585 |
| | | Final | 65.76% | 44.66% | 56.03% | 56.45% |
| | T15 | Precision | 0.7705 | 0.5973 | 0.5623 | 0.7891 |
| | | Recall | 0.5685 | 0.5973 | 0.5623 | 0.7891 |
| | | F1 Score | 0.6543 | 0.5973 | 0.5623 | 0.7891 |
| | | Final | 56.85% | 33.96% | 31.97% | 44.86% |

1. ***Training 1: Regression Model Training***
   The training data is introduced into a regression-based model, which undergoes training for 10 epochs per individual dataset, culminating in a cumulative total of 90 epochs across all datasets. Due to the similarity of the SCADA data and the substantial amount of data-logs over 6 years, there was not much need for a huge number of epochs.

2. ***Testing 1: Regression Model Testing***
   After model training, the testing data set was used to forecast incoming alarms, producing a binary dataset with '1' for alarm presence and '0' for normal operation. This output is used for input compilation in the Classifier step.

3. ***Input-Output Data Compilation for Classifier-based Training***
   The regression-model-based binary outputs corresponding to '1' are separated into separate datasets with their corresponding input unit windows. This gives us a time-series dataset of discrete input units comprising only alarm-associated behavior. This new input dataset is consistent with discrete units where the time-step distance among two adjacent logs can vary depending on the alarm occurrence in the original dataset. This kind of pattern is not usually associated with time-series where there is uniformity among the time-steps across the entire dataset length. The persistent characterization of this data as a time series remains, primarily due to its derivation from the original time-series dataset utilized by the regression model, as well as the fact that the successive input units continue to be organized in the sequence of their original temporal occurrence. Of-course, it must be underscored that the actual ordering of the time series is inconsequential in this context.

   WT alarms frequently occur in clusters, where one alarm triggers others. While some studies group clustered alarms into unique codes to analyze fault behavior, our focus on predicting incoming alarms requires only temporal occurrence data from the time series, making such clustering unnecessary for this application In addition to this, prior studies focus on false alarms in WT fault analysis; our work targets alarm prediction rather than fault behavior assessment. However, regression errors propagate through our two-step methodology: (1) The LSTM model processes 2-hour SCADA windows to forecast alarms (e.g., 7/10 predictions indicate alarms, with 1 false positive); (2) The classifier then processes all predicted alarms (including errors), introducing additional inaccuracies (e.g., correctly classifying only 5/6 true alarms). Thus, the final accuracy (5/7 here) must account for both regression false positives and classification errors through a correction step.

4. ***Training 2: Classifier-based Training***
   As mentioned in , the classification step includes 3 individual models, and for each model, an independent model training was performed on the only-alarm-based dataset we compiled in Section 3.

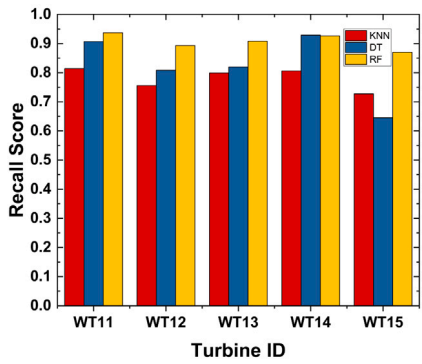

(a) FW: 10-minutes

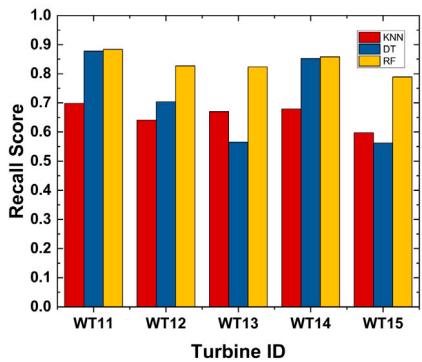

(b) FW: 20-minutes

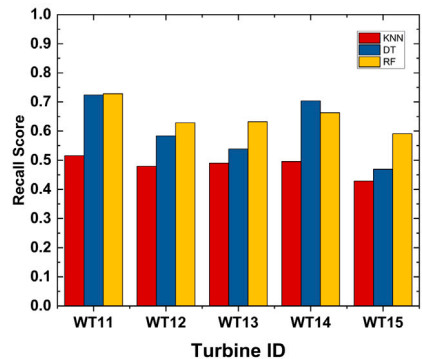

(c) FW: 30-minutes

**Fig. 9.** Recall Score—Classification Models results.

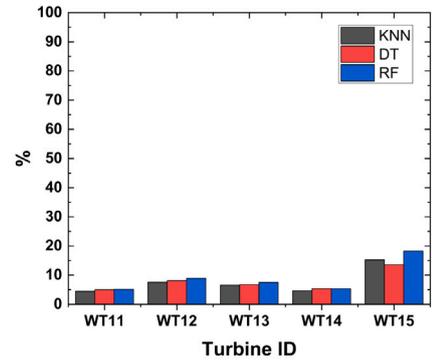

(a) FW: 10 minutes

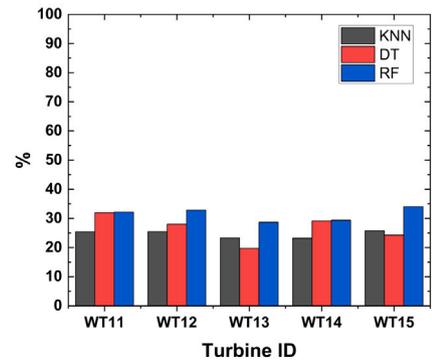

(b) FW: 20 minutes

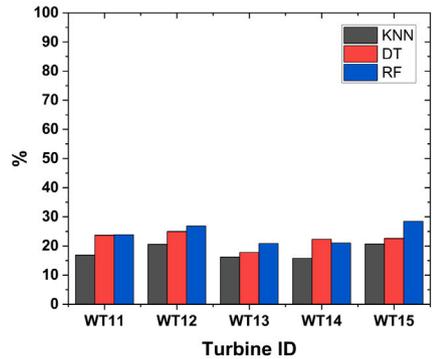

(c) FW: 30 minutes

**Fig. 10.** FPAF percentage.

5. *Testing 2: Classifier-based Testing*

In this subsequent section, the classification testing phase was performed. The step design follows such that for each individual model, parallel model testing would occur, i.e., bagging, and whichever model would perform the best based on the recall score would be forwarding its output as the final output.

The AFC methodology outputs classified alarms with 10–30 min advance warnings through its specialized two-stage architecture. By decoupling and optimizing the regression (forecasting) and classification tasks independently, AFC achieves superior performance compared to conventional approaches. This represents a paradigm shift from traditional risk-assessment methods that simply classify alarms in real-time to identify emerging faults - a reactive strategy that begins only after fault initiation. While existing studies ( [53–55]) have focused on early fault detection, AFC uniquely prioritizes preemptive alarm detection and aversion. Since faults typically emerge from accumulating minor alarms and warnings, our approach intervenes earlier in the failure progression chain. By forecasting and preventing constituent alarms rather than detecting developing faults, AFC offers greater potential to arrest deterioration before serious damage occurs. This upstream intervention in the failure progression chain offers superior containment potential compared to traditional fault-based methodologies. The next section will present the results achieved from this.

4. **Results & discussion**

In the initial forecasting step of the AFC, experiments were carried out using three different FW sizes: 10 min, 20 min, and 30 min. Following the regression forecasting, the identified alarms undergo classification using the aforementioned classification models. The evaluation metrics for both regression and the classification steps are tabulated in Tables 2 and 3; the breakdown and the empirical extrapolation of these results are individually quantified and discussed in the succeeding section, concluding the consolidated results, AFC final output, in Section 4.5. We adopt **recall** as the primary performance metric since our forecasting task and subsequent binary classification of alarms yield inherently discrete outcomes. Recall—also known as the true-positive rate—directly measures the fraction of actual alarms that the model correctly identifies, which is crucial when overlooking an alarm can have serious consequences. We additionally report **precision** and the $F_1$-**score**—the harmonic mean of precision and recall—to gauge the trade-off between false positives and false negatives and to guide hyperparameter tuning in our Bayesian classifiers; although these metrics are not used to select the final model, they provide valuable insight into model calibration and the extent to which predictions deviate by one or two time-steps. We also compute **accuracy** (the overall percentage of correct forecasts) to give a global sense of performance across both alarm and no-alarm instances. Finally, we introduce a custom FPAF—defined in Section 4.3—which quantifies the proportion of non-alarm periods that the model erroneously flags, thereby complementing recall with a domain-specific penalty on spurious alerts.

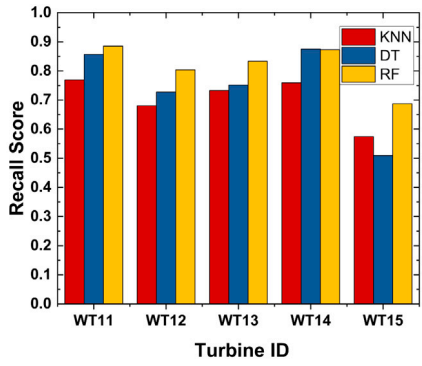

(a) FW: 10 minutes

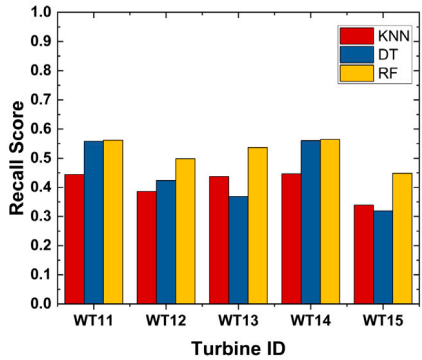

(b) FW: 20 minutes

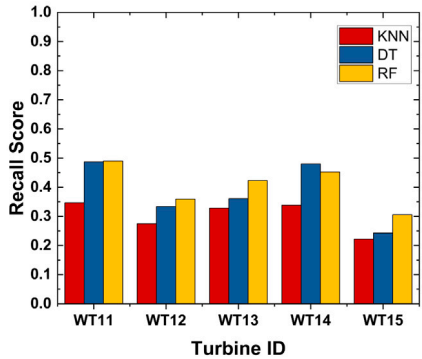

(c) FW: 30 minutes

**Fig. 11.** Final accuracy: post-FPAF correction.

**Table 3**
Evaluation metrics for different forecast windows (FW3) and turbines. The final score represents the post-FPAF removal.

| Forecast | Turbine | Evaluation parameter | LSTM | KNN | DT | RF |
|---|---|---|---|---|---|---|
| FW3 | WT11 | Precision | 0.4485 | 0.5154 | 0.7241 | 0.7285 |
| | | Recall | 0.6727 | 0.5154 | 0.7241 | 0.7285 |
| | | F1 Score | 0.5382 | 0.5154 | 0.7241 | 0.7285 |
| | | Final | 67.27% | 34.67% | 48.71% | 49.00% |
| | WT12 | Precision | 0.7631 | 0.4797 | 0.5835 | 0.6284 |
| | | Recall | 0.5719 | 0.4797 | 0.5835 | 0.6284 |
| | | F1 Score | 0.6538 | 0.4797 | 0.5835 | 0.6284 |
| | | Final | 57.19% | 27.44% | 33.37% | 35.94% |
| | WT13 | Precision | 0.5009 | 0.4903 | 0.5387 | 0.6320 |
| | | Recall | 0.6700 | 0.4903 | 0.5387 | 0.6320 |
| | | F1 Score | 0.5732 | 0.4903 | 0.5387 | 0.6320 |
| | | Final | 67.00% | 32.85% | 36.09% | 42.34% |
| | WT14 | Precision | 0.4962 | 0.4961 | 0.7036 | 0.6628 |
| | | Recall | 0.6825 | 0.4961 | 0.7036 | 0.6628 |
| | | F1 Score | 0.5746 | 0.4961 | 0.7036 | 0.6628 |
| | | Final | 68.25% | 33.86% | 48.02% | 45.24% |
| | T15 | Precision | 0.7453 | 0.4289 | 0.4690 | 0.5910 |
| | | Recall | 0.5175 | 0.4289 | 0.4690 | 0.5910 |
| | | F1 Score | 0.6109 | 0.4289 | 0.4690 | 0.5910 |
| | | Final | 51.75% | 22.20% | 24.27% | 30.59% |

$$\text{Accuracy} = \frac{TP + TN}{TP + TN + FP + FN}$$

$$\text{Precision} = \frac{TP}{TP + FP}$$

$$\text{Recall (True Positive Rate)} = \frac{TP}{TP + FN}$$

$$F_1\text{-score} = 2 \cdot \frac{\text{Precision} \times \text{Recall}}{\text{Precision} + \text{Recall}} = \frac{2TP}{2TP + FP + FN}$$

Fig. 8 delineates the sequential methodology involved in evaluating AFC accuracy. The identification of advanced alarms within the regression step is assessed by isolating true and false positives, providing a quantitative basis for evaluating the regression model's performance. Since this phase integrates FW, serving as the critical inflexion point in forecasting, it simultaneously imposes constraints on extending forecasting horizons beyond 30 min. Utilizing the binary alarm occurrence forecast ('1' identifier from Fig. 1), only the time windows corresponding to alarm events are inputted into the classification phase, where a tri-model parallel classification framework is employed to assign nomenclature to the alarms. Subsequently, outputs are processed through the bagging algorithm to derive the optimal model. Given the sequential pipelining of the AFC framework, false positives originating from the regression phase propagate into the downstream classification phase. Therefore, their systematic elimination is imperative for accurately assessing the efficacy of the final bagged results. Notably, the principal peripety is observed during the regression phase, where classification step predictions exhibit a heteroscedastic pattern across forecasting intervals from 10 to 30 min. However, upon excising FPAF (false positive alarm forecasts) from the classification phase, the model accuracy demonstrates a pronounced inflexion, as illustrated in Fig. 14.

*4.1. Regression step results*

The recall score was chosen as the evaluation criteria for accurately forecasting incoming alarms for a given FW; results are presented in suggesting more predictable or less noisy alarm data. The model's performance is strongest for the shortest FW1 but faces challenges in longer-term predictions. Further optimizations regarding data quality might be required to enhance its performance for extended FW.

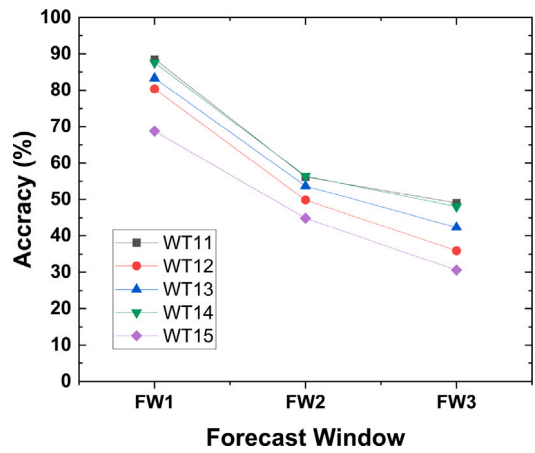

**Fig. 12.** Final output accuracy based on bagging operation.

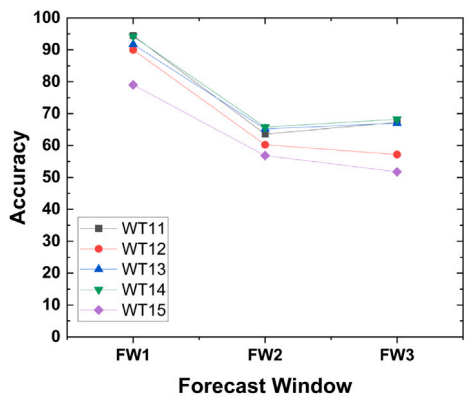

(a) LSTM—regression model

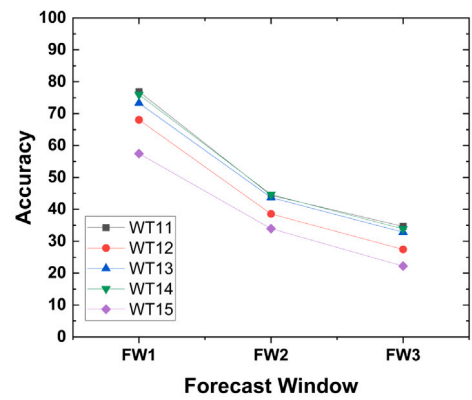

(b) KNN

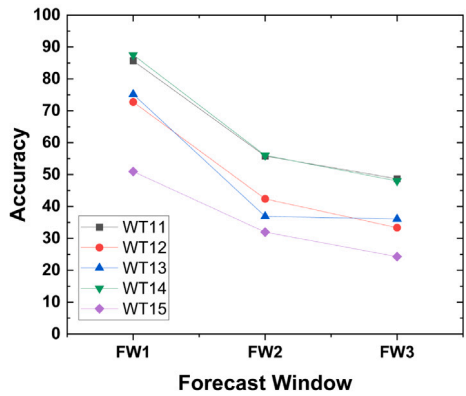

(c) DT

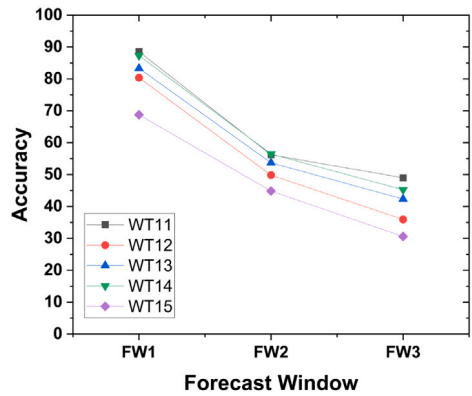

(d) RF

**Fig. 13.** Effect of FW on accuracy—the graphs show the decrease in the accuracy for all models as the FW increases: (a) LSTM-based regression model, (b) KNN, (c) DT, and (d) RF.

Fig. 7. The model's accuracy decreases significantly as FW increases from 10 min to 30 min, limiting its expansion further. Other parameters, like precision, root mean Squared Error (RMSE), are often preferred for regression-based models; however, the binary nature of the regression model makes the recall score a much more preferred option. Results showed high accuracy across all turbines, with WT11 and WT15 showing the highest accuracy. However, as the FW window grew, the accuracy dropped, indicating potential challenges in long-term forecasting. Turbines WT14 consistently showed higher accuracy,

*4.2. Classification step results*

Fig. 9 presents the classification outcomes for each model. A comparison of models in the AFC framework shows that RF is always more accurate than the others across all FWs and turbines. While DT generally ranks second, performing better than KNN trailing RF, KNN consistently exhibits the lowest accuracy. As FWs increases, all models experience a decline in accuracy, highlighting the difficulties of long-term prediction. Notably, turbine WT11 often achieves the highest accuracy, suggesting its alarm data may be more predictable. Overall, RF proves to be the most resilient model for the AFC approach.

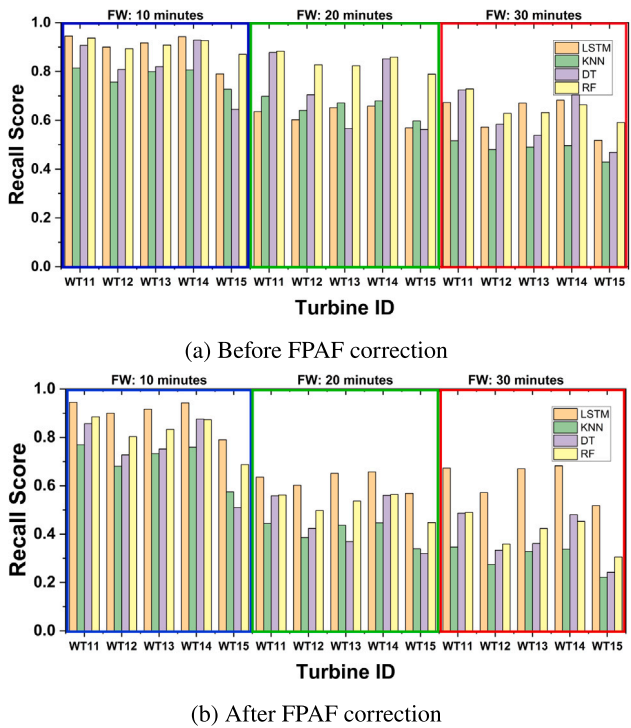

**Fig. 14.** FPAF correction effect: Graph (a) shows the recall score before the FPAF correction, and Graph (b) shows the recall score after the FPAF correction. A significant decrease in the recall score is observed from pre- to post-alarm correction.

### 4.3. FPAF effect

The alarm classification results from Section 4.2 do not represent the definitive accuracy of the AFC approach, as the accuracy from the regression step has not been factored into this. To get accurate alarm classification, the FPAF quantity (or what should be called missed alarms) is subtracted from the accuracy output of the classification step. Fig. 10 graphs represent the FPAF quantity for each turbine. The FPAF rates generally increase with the FW size across all models and turbines, indicating the increasing complexity of longer-term predictions. KNN generally exhibits lower FPAF rates compared to DT and RF, especially for longer FWs.

While methods such as uncertainty-aware Monte-Carlo Dropout have been shown to provide confidence bounds on LSTM forecasts and reduce false alarms in SCADA systems [56] and targeted preprocessing has cut false-positive rates in wind-turbine fault detection [39], further gains might be realized by end-to-end multi-task learning that jointly optimizes forecasting and classification layers [57] or by incorporating feedback loops that re-evaluate borderline predictions via temporal smoothing. We deliberately refrained from applying these extensions here in order to preserve a clean, sequential benchmark of AFC's two-stage design and because their practical success typically hinges on detailed alarm-specific calibration—manual priority settings, varying criticality levels, and bespoke response protocols that differ across turbine models and operators [58]. Consequently, we present these strategies as future research avenues, empowering practitioners to adapt and validate them within their own operational and data-quality constraints, without compromising the transparent assessment of each AFC component.

There is another unintended consequence of FPAF that needs to be pointed out. For this, we are to refer to two specific figures, Fig. 7 (forecast results from regression) and Fig. 10 (FPAF effect for the same forecasts). The key point to note here is that there is a relative proportional drop in forecast results as the window size is increased. However, in Fig. 10, there is a significant jump in the FPAF for the 20 min case, and then it goes down again for 30 min. This could better be referred to as a spike. The classification results (i.e., Fig. 9) are relatively irrelevant in this situation as the FPAF quantity is derived from the missed alarms during forecasting, which was only done in the regression step. Upon investigating the reasoning behind this behavior, we came to the realization that this is based on time proximity. Simply put, in a 20 min window, some alarms hold a relatively varying time-domain relation to the earlier turbine behavior. This means in the 20 min window, there is always a good chance for a specific alarm to occur, but for the 30 min window, since the timeline is significantly far enough that the model can relatively assure that there is an incoming alarm or not, thus leading to less FPAF quantity. Therefore, the 20 min window becomes the inflexion point where the model remains relatively unsure about an incoming alarm, leading to high false predictions. When compensated for these false predictions, we see the proportional drop in accuracy from Figs. 9(b) to 10(b). This gives us an empirical reasoning on why the window size beyond 20 min shows very diminishing correlation to alarms; therefore, extending the window further proves counterproductive. It can be unequivocally stated that alarm data could only be utilized in a succinct and short-term manner, ruling out options of much larger windows of days, months, or even years. Of course, there have been multiple studies that have definitively extracted the health status of a turbine weeks or months ahead. But that is more of a probabilistic approach where the task is limited to assessing whether the turbine would fail in future periods or not based on alarm history. The types of alarms and their temporal distribution, i.e, what alarm and when they would occur, have mostly been ignored. The reasoning behind being the FPAF effect that we described— limiting FWs up to 20–30 min.

### 4.4. True classification step accuracy

The true accuracy AFC can be determined by subtracting the total number of correctly forecasted and classified errors from the total number of actual alarms that occurred in the initial dataset. This calculation provides a direct measure of the accuracy of fault classification without considering false positives or false negatives, i.e., false alarms. This metric is illustrated in Fig. 11; across all FWs, RF outperforms KNN and DT, showing higher accuracy, especially for WT11, while WT15 consistently exhibits the lowest accuracy across all models and FW, key points are enumerated as follows:

1. Overall Model Performance: RF consistently outperforms the other models across all FWs and turbines, reflecting its robustness in handling the data's complexity.
2. Impact of FPAF Adjustments: Adjusting for false alarms has generally decreased the accuracy values across all models, emphasizing the importance of accounting for misidentifications.
3. Performance by Turbine: WT15 consistently shows the lowest accuracy across all models and FWs, indicating challenges in predicting alarms for this turbine.
4. FW Impact: As FW size increases, the accuracy tends to decrease across all models, highlighting the increasing complexity of longer-term predictions.
5. Model Suitability: RF remains the most reliable model, offering the highest accuracy across various FWs and turbines.

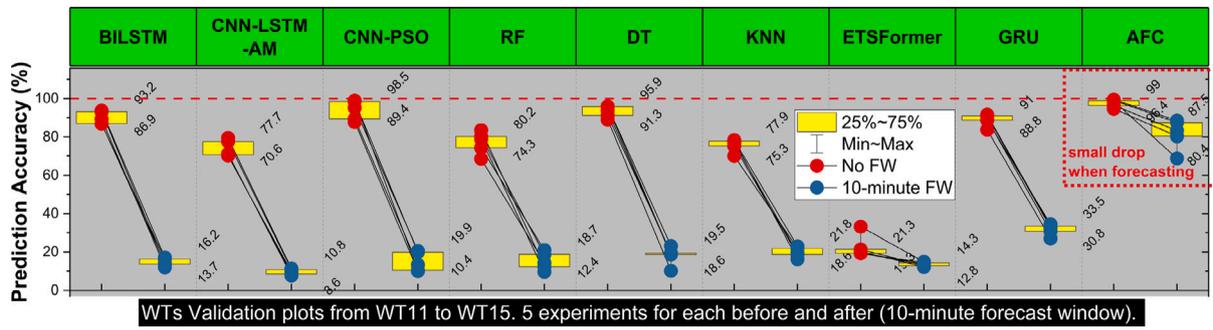

**Fig. 15.** Average accuracy comparison of forecasting vs. classification across nine models, with data labels shown on bars.

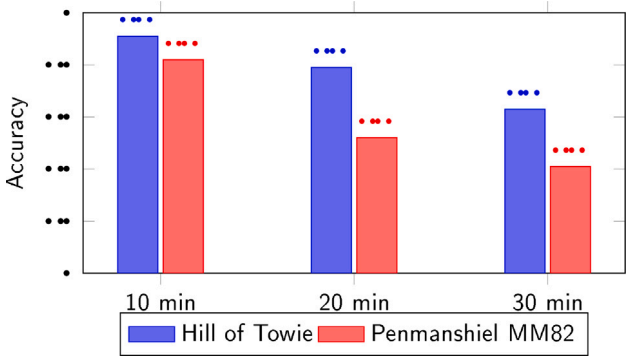

**Fig. 16.** Forecasting accuracy comparison across datasets and forecasting windows.

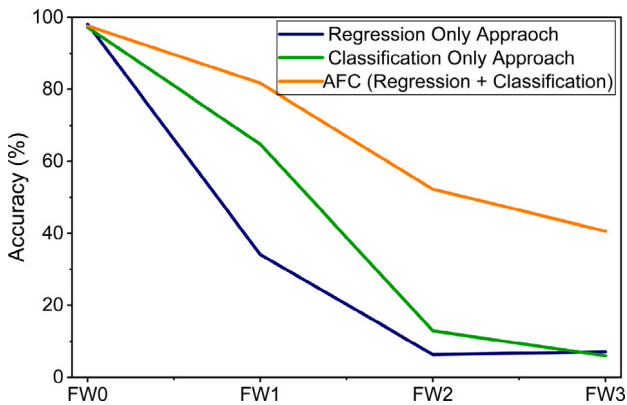

**Fig. 17.** Standalone Regression and Classification Results. Forecasting window sweeps from FW0 (no forecasting) to FW3 (30 min).

6. Accounting for FPAFs: FPAFs have a noticeable impact on the accuracy of alarm predictions. Properly accounting for these misidentifications is crucial for improving the reliability of the forecasting system.
7. Further Considerations: Continuous refinement and evaluation of the models, especially focusing on improving predictions for challenging turbines like WT15, are essential for enhancing the system's overall performance.

Before moving on to the next phase, although not aligned with the context, a certain aspect of accuracy, besides the FPAF-refined, needs to be discussed. As earlier explained, the dataset is consistent with 223 unique alarms; the accuracy discussed so far is a cumulative accuracy equated by accounting for all the initially triggered alarms, irrespective of their alarm codes, and segregating the total instances that were accurately forecasted. The subjective accuracy of each alarm type is sidelined. However, if concaved up to individual alarms, the accuracy parameters portray an unparalleled picture; Figs. 21, 22, and 23 portray the accuracy metric for each unique alarm; alarms abundant in frequency circumvent an imposition on entire model learning, biasing towards fidelity for this more frequent alarm and providing them with invigorated prediction accuracy while sidelining the less frequent ones. This does mean that the alarms that are non-frequent might get missed; on the contrary however, as a blessing in disguise, since alarms with a high proportion of occurrence are predicted with a similar higher congruence degree of accuracy, this implies the overall accuracy will also be skewed to provide a better prediction accuracy, thus avoiding the majority number of alarms from happening beforehand, proving its empirical efficacy. To counteract this, data-level methods such as random oversampling or synthetic techniques like SMOTE and ADASYN have been shown to boost minority-class sensitivity in real WT SCADA datasets. At the algorithm level, class-weighted loss functions or focal loss can rebalance learning by penalizing misclassification of infrequent alarms more heavily [59]; focal loss notably improved $F_i$ scores by 4%–6% in blade-icing prediction tasks [60]. By either amplifying scarce alarm samples or reshaping the loss landscape, these strategies help the classifier form more equitable decision boundaries across all alarm types. Integrating oversampling with a weighted or focal-loss objective thus offers a well-founded path to mitigate performance disparity among alarm classes in future AFC deployments.

Appendix gives the individual alarm accuracy, while Figs. 24, 25, and 26 in the appendix detail the miss-identification parameter by displaying the heat map for each experiment. The alarm frequency is non-uniformly distributed, with certain alarms being present a few orders higher in magnitude. This denomination skews the training process, funneling the prediction capabilities in favor of those with higher frequency, resulting in sparsely occurring alarms being wholly ignored. However, since alarms forecast with higher frequency resulted in exceptional performance, uplifting the total aggregate AFC provides a solution to the majority of alarms. Table 4 provides the contingency matrix of the future classification results; the False Positives (FP) are the falsely predicted alarms (inaccurate tag or false alarm when there was suppose to be none), False Negatives (FN) are the escaped alarms, and True Positives (TP) are the accurately predicted alarms. RF overall performance is again shown as superior in the table (see Table 5).

### 4.5. Final output

The bagging technique selects the best-performing model for each FW, simplifying the decision-making process. The bagging technique effectively leverages the strengths of each model, resulting in higher overall accuracy. Fig. 12 illustrates the performance of the final result, showcasing the percentage of accurately forecasted alarms; the tabulated summary is showcased in Table 5. The results presented in this section provide valuable insights into WT labeled-alarm forecasting, highlighting the potential of ML models in enhancing system reliability and efficiency. While significant progress has been made, there remain challenges and opportunities for future exploration and refinement. By continuing to innovate and adapt, we can pave the way for a more sustainable and efficient wind energy sector. Let us enumerate some of the key insights:

1. Forecasting Window Effect: The AFC methodology's accuracy is affected by the increasing FW. The general trend shows a sharp decline from 10 min to 20 min, but moving from 20 min to 30 min does not significantly decrease accuracy. However, increasing FW would render the methodology impractical. For 30 min of FW, the accuracy ranges between 30 and 50%; see Fig. 12. While this accuracy might be useful for some applications, further increase would be counterproductive.

2. FPAF Effect: The accuracy and viability of the AFC methodology are influenced by the quantity of FPAFs. As the FW size

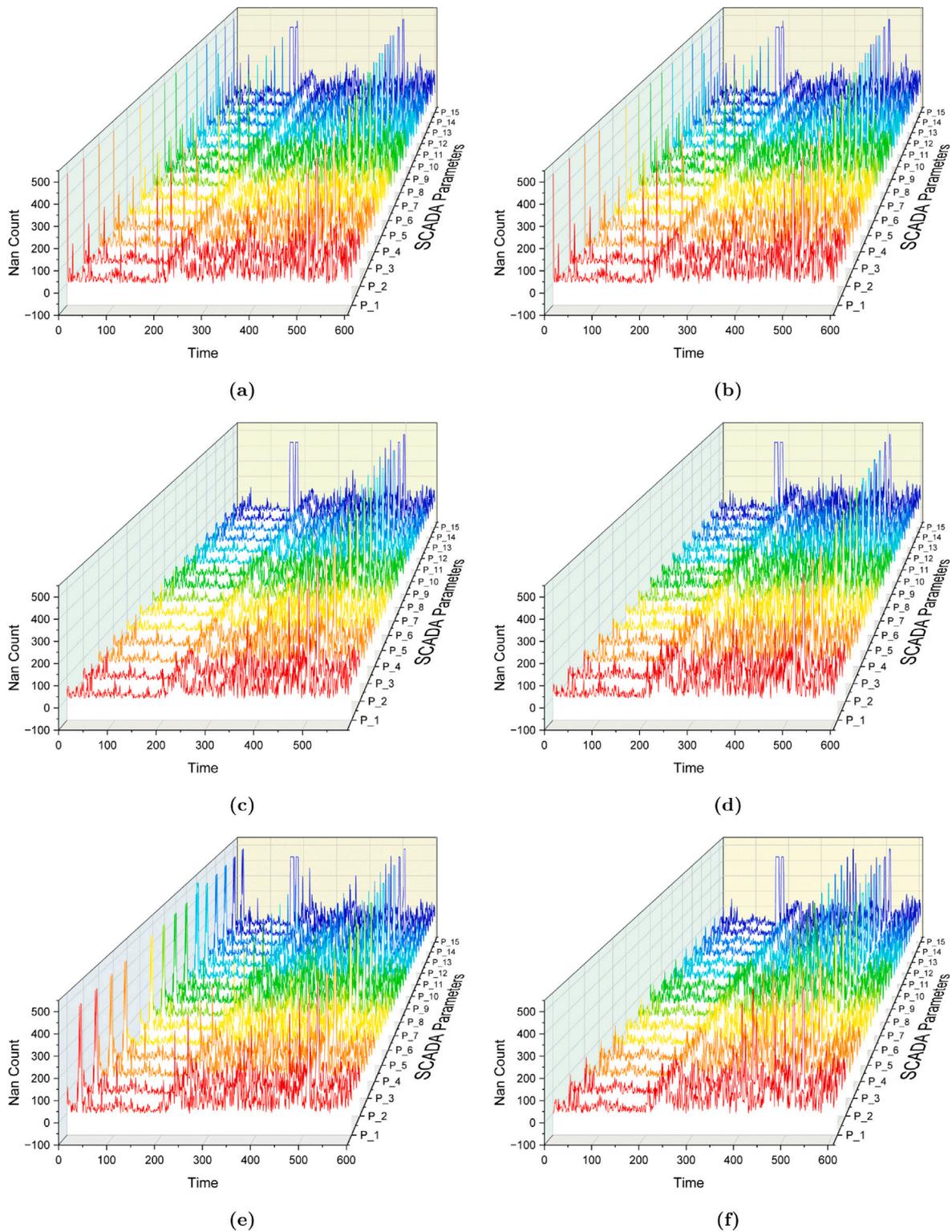

**Fig. 18.** Time-series NaN values distribution of first 15 parameters (a) WT01 (b) WT02 (C) WT04 (d) WT05 (e) WT06 (f) WT07.

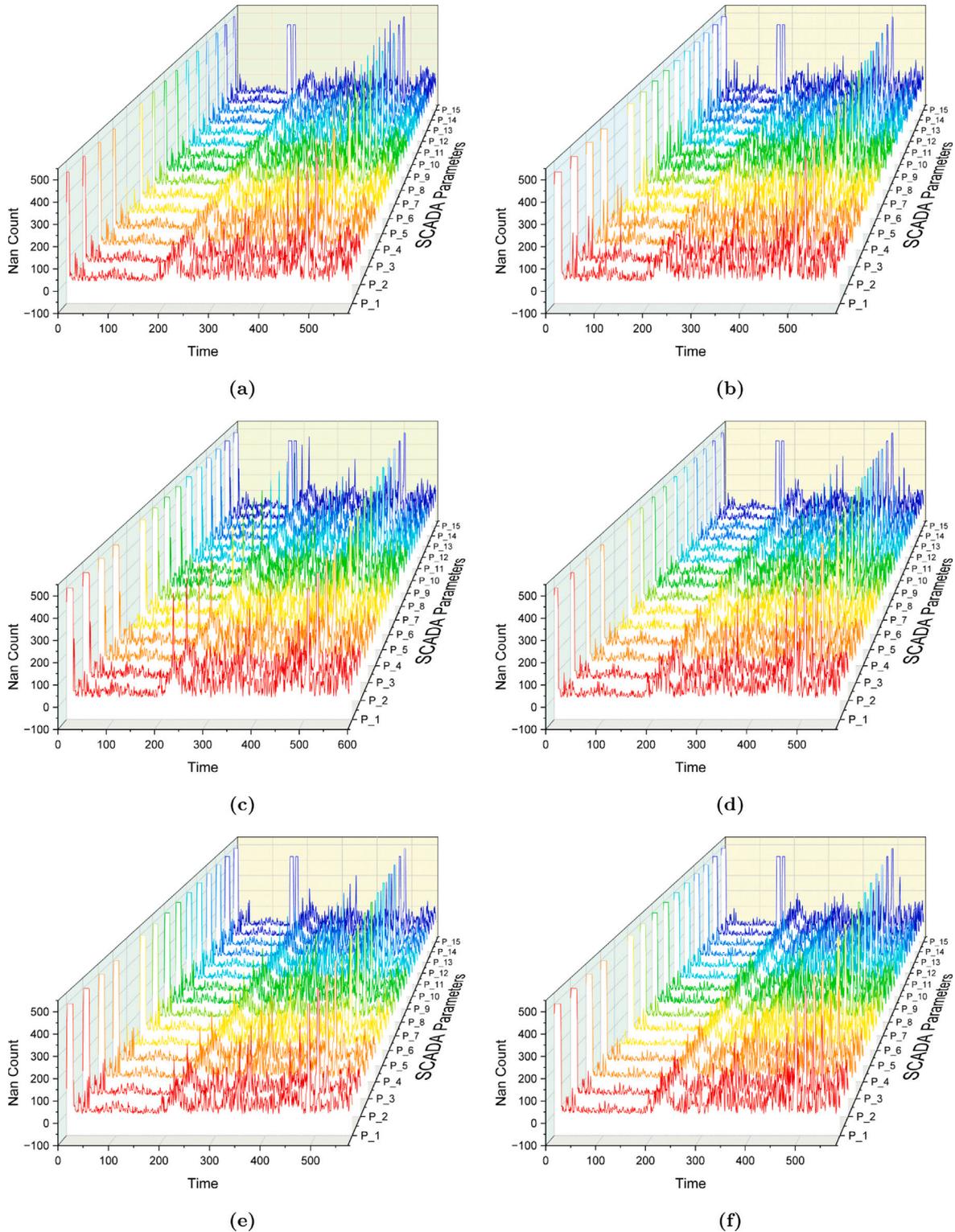

**Fig. 19.** Time-series NaN values distribution of first 15 parameters (a) WT08 (b) WT09 (C) WT10 (d) WT11 (e) WT12 (f) WT13.

increases, the accuracy decreases, as shown in Fig. 14. The final accuracy of the methodology is dependent on two key steps: regression-based forecasting accuracy and classifier-based classification accuracy. The LSTM regression-based model's accuracy remains unaffected, while the recall score for the LSTM-regression model shows inaccuracies.

3. Bagging Technique Efficacy: The bagging technique enhances overall accuracy in alarm classification by selecting the best-performing model for each FW. The RF model is dominant, but bagging techniques should be preferred due to variability in performance across scenarios. The labyrinth of ML models offers opportunities for integrating new models if less noisy and refined SCADA data is available.

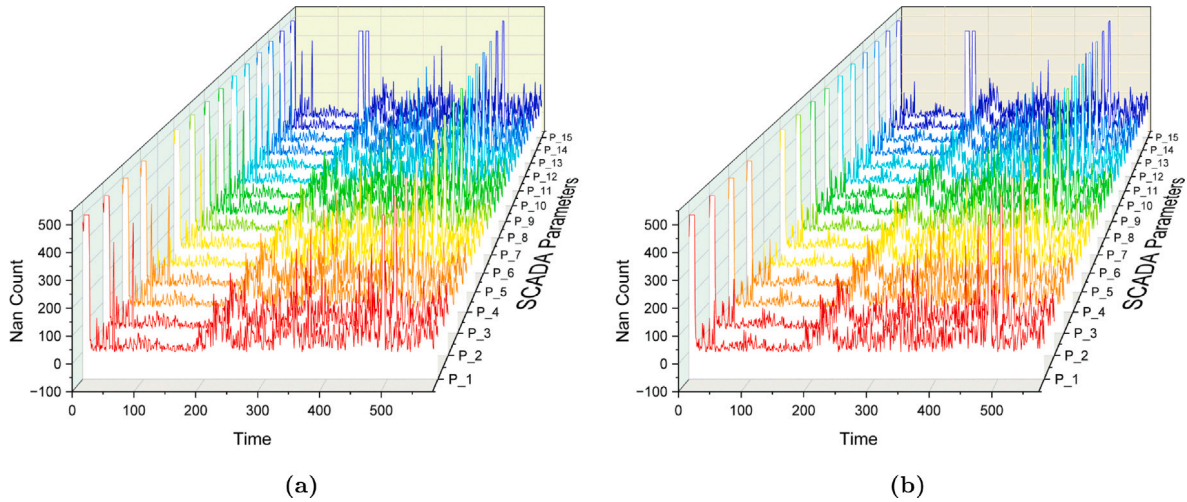

**Fig. 20.** Time-series NaN values distribution of first 15 parameters (a) WT14 (b) WT15.

**Table 4**
Contingency statistics per forecast window (averaged over T11–T15).

| (a) FW1 | | | |
|---|---|---|---|
| Model | FP | FN | TP |
| $KNN_2$ | 0.077 | 0.219 | 0.703 |
| $DT_2$ | 0.078 | 0.178 | 0.744 |
| $RF_2$ | 0.090 | 0.093 | 0.837 |

| (b) FW2 | | | |
|---|---|---|---|
| Model | FP | FN | TP |
| $KNN_2$ | 0.246 | 0.343 | 0.411 |
| $DT_2$ | 0.266 | 0.288 | 0.446 |
| $RF_2$ | 0.314 | 0.164 | 0.522 |

| (c) FW3 | | | |
|---|---|---|---|
| Model | FP | FN | TP |
| $KNN_2$ | 0.178 | 0.518 | 0.302 |
| $DT_2$ | 0.223 | 0.396 | 0.381 |
| $RF_2$ | 0.242 | 0.352 | 0.406 |

**Table 5**
Final output accuracy based on bagging operation (corresponding to Fig. 12).

| FW | WT11 | WT12 | WT13 | WT14 | WT15 | Average |
|---|---|---|---|---|---|---|
| FW1 | 0.89 | 0.80 | 0.83 | 0.88 | 0.69 | 0.82 |
| FW2 | 0.56 | 0.50 | 0.54 | 0.56 | 0.45 | 0.52 |
| FW3 | 0.49 | 0.36 | 0.42 | 0.48 | 0.31 | 0.41 |

*4.6. Validation*

To validate the effectiveness of the proposed AFC approach, we conducted a comprehensive comparison against several state-of-the-art models widely adopted by other researchers for similar tasks. These models include bidirectional long short-Term memory (BiL-STM) [61], convolutional neural network (CNN)–LSTM with attention mechanism (AM) (CNN-LSTM AM) [29,62], standalone CNN with particle swarm optimization particle swarm optimization (PSO)(CNN-PSO) [63], RF [64], DT [41], KNN [65], transformer-based architecture ETSFormer [66], and gated recurrent unit (GRU) [67]. The evaluation was performed across all test turbines, with and without a forecasting window (10 min), to ensure robustness and consistency. Fig. 15 illustrates the validation experiments results. For each given test dataset (5 in total: WT11 to WT15), first, the experiments were tested on the simple classification task, i.e., simply finding out the tags for a given alarm. Most validation models performed respectably for the simple prediction task with no FW, except ETSFormer. However, the contrast difference was significantly apparent as the forecasting task was applied in the second trial; a window of 10 min forecast caused excessive attenuation to prediction accuracy. Results demonstrate that AFC consistently outperforms these baseline models in terms of accuracy, particularly under the 10 min forecasting scenario, highlighting its superior capability in handling alarm classification under real-world operational constraints. By decoupling forecasting and classification, AFC leveraged a forecasting network optimized solely for future prediction (LSTM), followed by an alarm classifier that ingests forecasted attributes in their most informative form. This specialization yields a forecasting-window accuracy of **81.696%**, far surpassing all benchmarks (next best: GRU at 34.5% for WT15); detailed results shown in Table 6. Even for the no-window classification accuracy, due to its divide-and-conquer stratagy, it outperformed (best result: **99.41%** for WT11) even state-of-the-art models (BiLSTM at 93.66%, CNN-PSO at 98.90%).

To attest the generalizability and robustness of the proposed AFC framework, we conducted a cross-site validation using the Hill of Towie Wind Farm Open Dataset recently released by RES and TRIG [68]. This comprehensive dataset comprises over eight years (2016–2024) of 10 min SCADA statistics, alarm logs, turbine metadata, and downtime information from 21 Siemens SWT-2.3-VS-82 turbines located in Scotland. The dataset has been curated specifically for research purposes and is exceptionally well-maintained, offering a high signal quality with almost negligible missing values (NaN), thus serving as an ideal benchmark for testing the forecasting capability of data-driven models. We fed this dataset through our AFC framework with no architectural modification and minimal data preprocessing and performed the same forecasting tasks for FW: 10, 20, and 30 min. This trial demonstrated significantly improved forecasting performance, achieving 0.91, 0.79, and 0.63 accuracy for the respective FWs, which outperforms the results on our original Penmanshiel MM82 dataset (0.82, 0.52, and 0.41), see Fig. 16. However, several caveats must be considered while interpreting these results. Firstly, the Hill of Towie dataset contains only 85 unique alarm codes, significantly less than the 221 codes present in the original dataset. This reduction in class diversity inherently simplifies the classification task. Secondly, the overall data quality of this external dataset is substantially higher, with well-structured alarm logs and nearly no data loss due to FPAF, a significant challenge in real-world SCADA operations. Moreover, the availability of over 300 SCADA parameters allowed for the selection of a richer and cleaner feature set ( 100 variables used), likely contributing to higher predictive accuracy. These differences highlight that while this validation experiment confirms the adaptability and strong performance of AFC under relatively ideal conditions, the more modest results obtained from our primary dataset still better reflect the framework's real-world applicability in industrial wind farm environments where data noise, gaps, and alarm ambiguities are natural. Nonetheless, this validation further underscores AFC's capacity to generalize across turbine models (Senvion MM82 vs. Siemens SWT-2.3-VS-82) and sites, affirming its usefulness in practical deployment scenarios while emphasizing the critical role of dataset quality and alarm code diversity in forecasting outcomes.

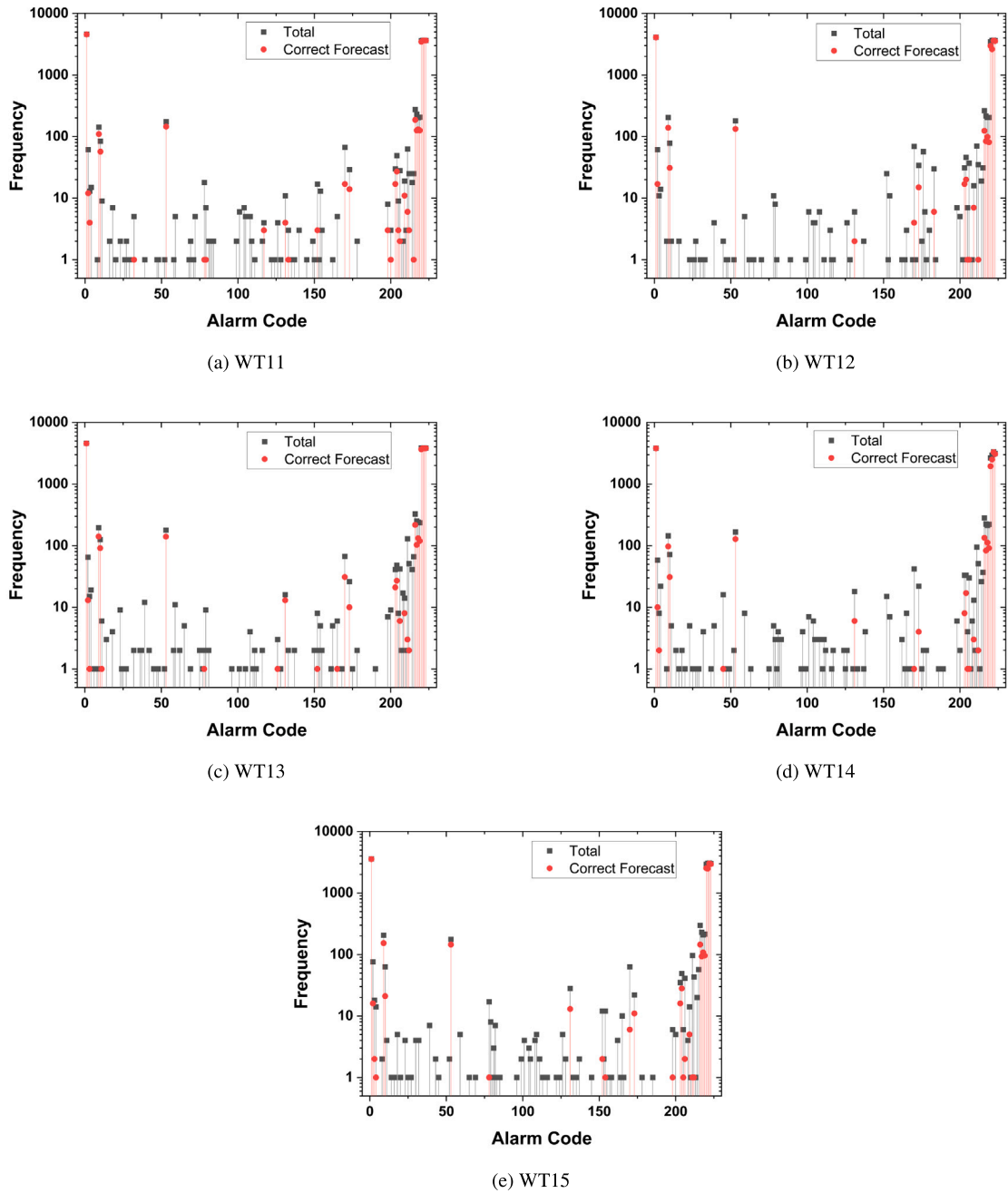

Fig. 21. Individual alarm forecast result—FW 10 minutes.

Table 6
Per-turbine results for both tasks (with and without forecasting window).

| Turbine | Model | | | | | | | | |
|---|---|---|---|---|---|---|---|---|---|
| | BiLSTM [61] | CNN–LSTM–AM [29,62] | CNN-PSO [63] | RF [64] | DT [41] | KNN [65] | ETSFormer [66] | GRU [67] | AFC$^{current*}$ |
| **No Forecasting Window** | | | | | | | | | |
| WT11 | 93.2 | 70.58 | 98.90 | 83.57 | 95.86 | 77.88 | 21.33 | 91.50 | 99.41 |
| WT12 | 86.85 | 79.41 | 89.41 | 68.53 | 89.02 | 70.09 | 19.25 | 83.72 | 94.70 |
| WT13 | 86.94 | 70.16 | 87.76 | 78.93 | 95.94 | 78.28 | 19.31 | 89.18 | 96.37 |
| WT14 | 89.39 | 71.18 | 98.51 | 80.20 | 93.80 | 75.25 | 20.08 | 90.99 | 98.96 |
| WT15 | 93.66 | 77.65 | 95.21 | 74.33 | 91.25 | 76.81 | 33.21 | 88.77 | 98.73 |
| **10 min Forecasting Window** | | | | | | | | | |
| WT11 | 13.67 | 8.60 | 10.36 | 12.36 | 18.75 | 16.25 | 12.83 | 33.48 | 88.53 |
| WT12 | 16.21 | 11.33 | 13.26 | 14.23 | 10.18 | 21.80 | 14.28 | 27.05 | 80.36 |
| WT13 | 14.21 | 9.60 | 10.05 | 9.65 | 23.11 | 23.00 | 13.51 | 30.78 | 83.31 |
| WT14 | 12.05 | 7.61 | 19.90 | 18.71 | 18.60 | 18.60 | 12.27 | 32.69 | 87.53 |
| WT15 | 17.11 | 10.80 | 20.50 | 21.01 | 19.46 | 19.40 | 14.80 | 34.50 | 68.75 |

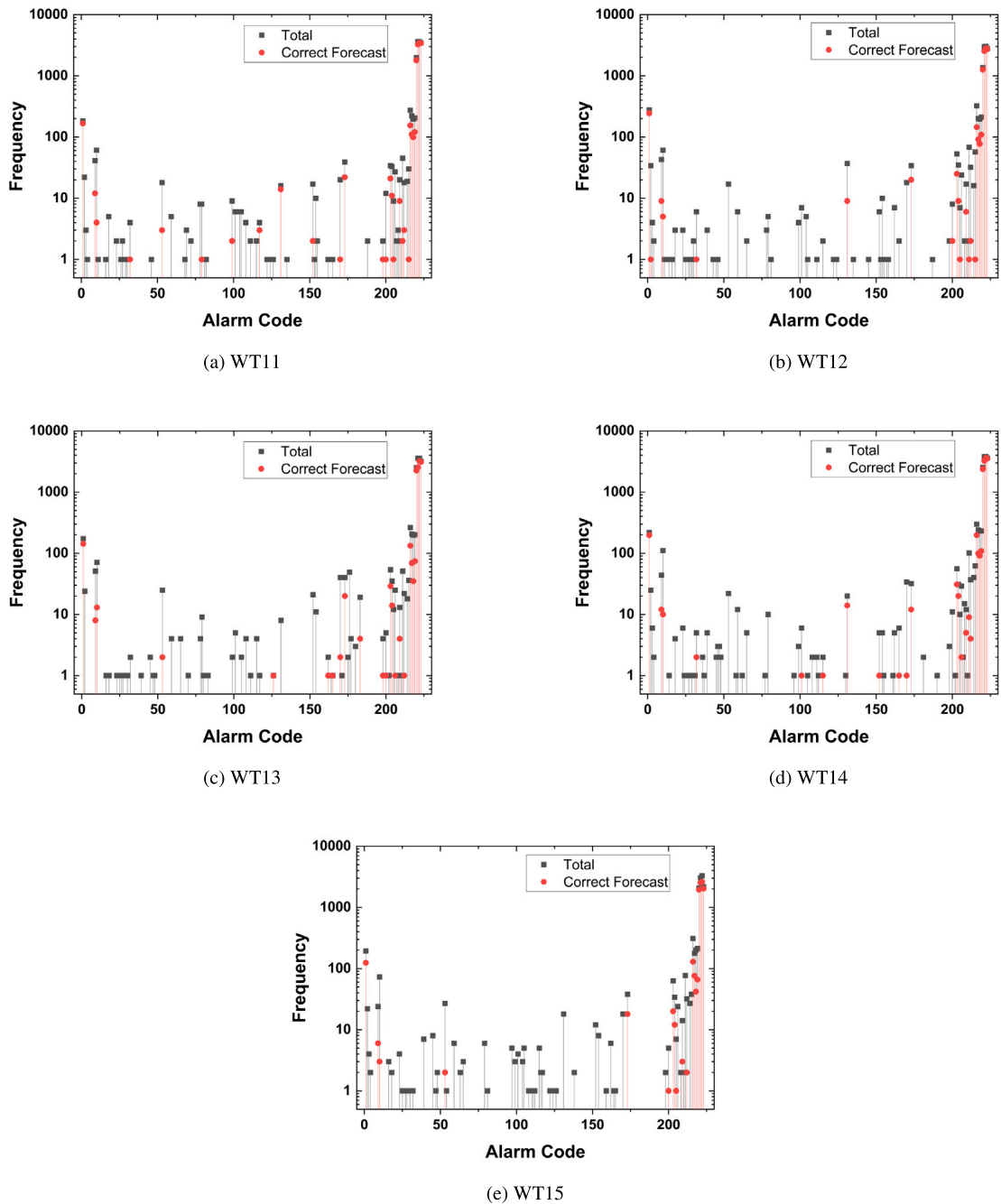

**Fig. 22.** Individual alarm forecast result—FW 20 minutes.

The key highlight that was conducive to successful alarm forecasting in AFC was the precarious design discussed in Section 2. By leveraging the pipelined architecture to perform series operations of regression and classification, we were able to achieve comparable results. To validate this hypothesis, the comparison result for alarm forecasting was done with these stand-alone approaches, presented in Fig. 17. All stated approaches performed as expected for the real-time prediction (FW0). However, as a FW was introduced, the performance took a steep plunge for the standalone approach. The regression model was excessively disappointing. Moreover, as the window size increases beyond 10 min, the results become almost fortuitous, and almost all temporal dependencies are missed by both regression and classification approaches. This further reinforces the hypothesis discussed earlier that the sheer number and the short-term nature of the wind turbine alarms make it very challenging to achieve it by conventional approaches. Therefore, the AFC architecture provides an overarching way that divides this single task for forecasting alarms into separate tasks of regression and challenging, thereafter deploying specialized models for each task to achieve unequivocally much optimal final results—sort of a modular approach based on divide and conquer.

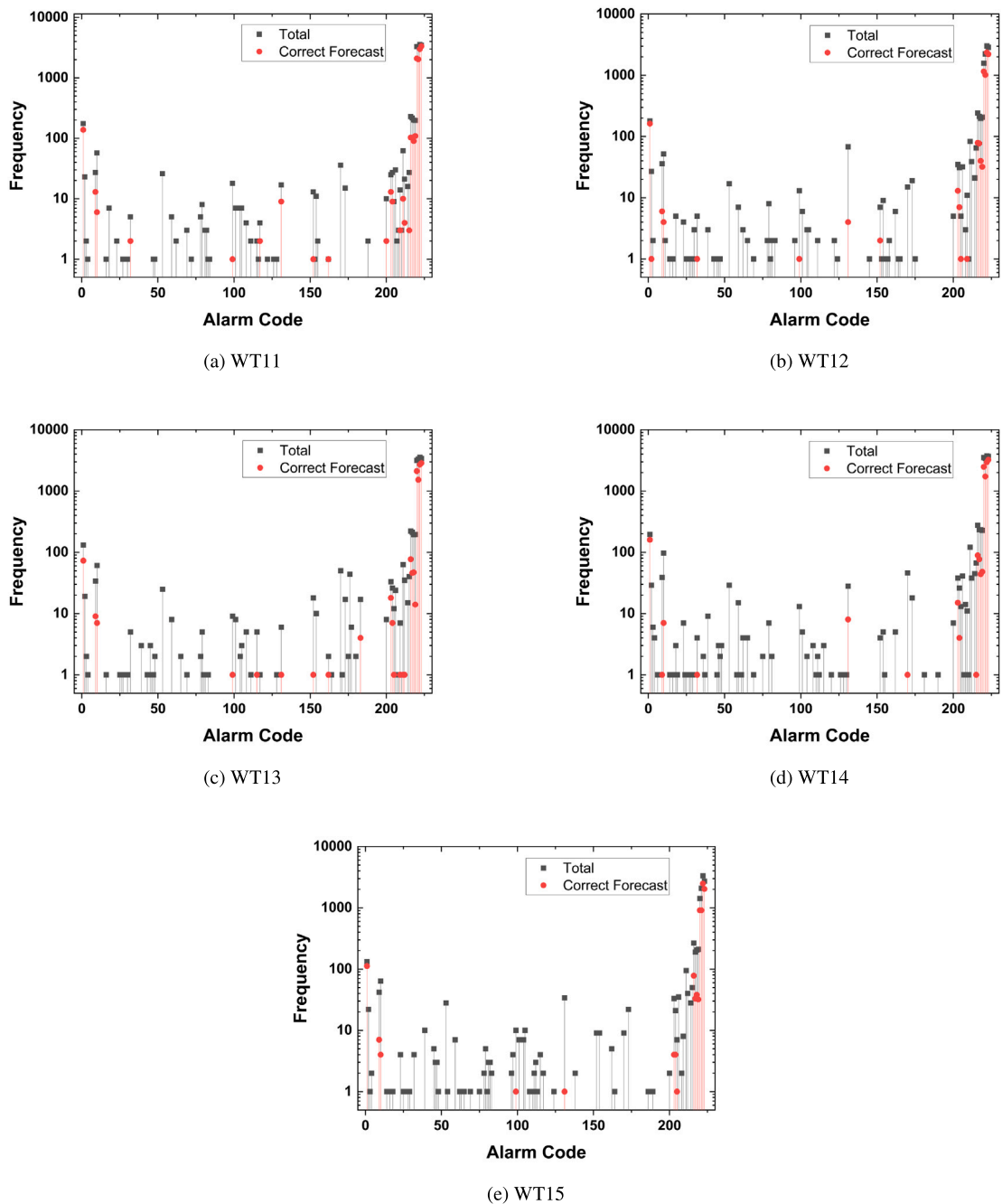

**Fig. 23.** Individual alarm forecast result—FW 30 minutes.

### 4.7. Implications and limitations

Accurate tagged alarm forecasting holds significant operational value, facilitating timely interventions, reducing downtime, and enhancing overall system efficiency for WTs. The results obtained from this study can potentially contribute to optimizing maintenance schedules and improving WT performance. However, due to the short FW size ranging from 10 to 30 min, rather than a physical response by a member(s) of the maintenance team, this approach should be much more beneficial if an automated response, either by the control system (which is much more preferred and recommended), is taken. On a side note, this should not preclude the possibility of a physical on-site maintenance response if needed; usually, maintenance personnel are present on-site all the time, so a response action from the control room could be considerable; however, this depends on the on-site call and the resources available. Moreover, since alarms are known in advance, and if correctly mitigated, this will address the much more persistent issue of FPAFs, thus adding to the resilience of the alarm-based conditional monitoring system.

The performance of ML models relies heavily on the quality and quantity of data used for training and validation. Data variability can impact predictive accuracy, especially in the wind industry, where companies often keep SCADA data confidential. Finding a better dataset with better refinement and less noise is challenging. Long-term predictions face challenges, and future research should focus on developing techniques to improve reliability. Alarm readings are short-term indicators, and extending the FW further will improve the control system response coordination platform.

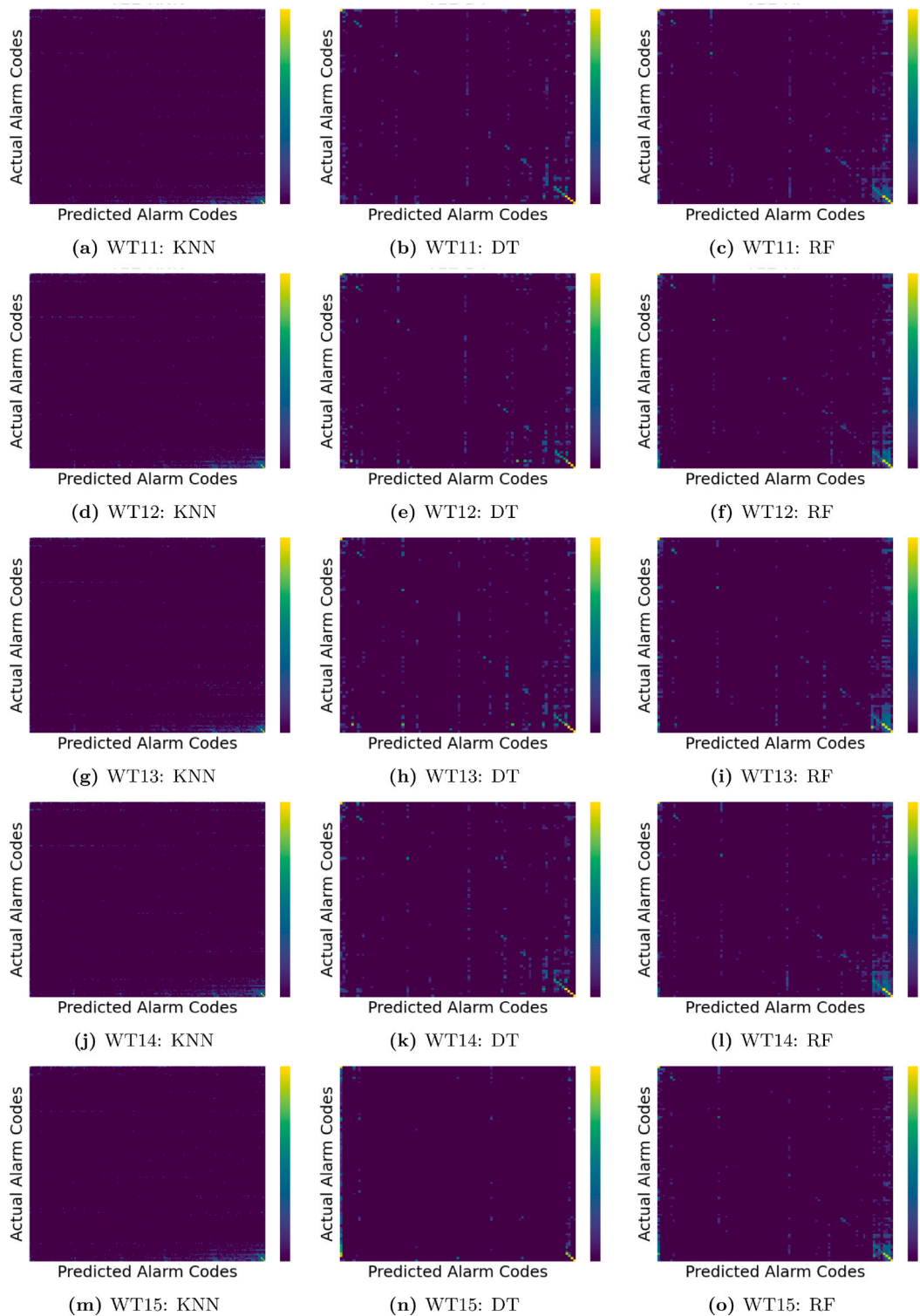

Fig. 24. FPAF (True positive—False negatives): FW 10 minutes.

## 5. Conclusion

1. The AFC methodology presented a novel approach to making use of the prowess elements from each regression and classification-based ML model, i.e., first successfully achieving the indication of alarms 10 to 30 min in advance by making use of the FW concept using LSTM regression models, then afterwards using classification models to find the relative code for those incoming alarms.

2. Using this approach, the study was able to successfully forecast alarms for a 10 min advance time window with the best accuracy of 88.5% for one of the WTs and 68.8% for the least accurate one. The average score is 82%, 52%, and 41% for the 10, 20, and 30 min forecast; the general downward trend as the FW size increases from 10 to 30 min is capped by further expansion beyond 30 min. AFC also superseded all other benchmark models not only for its niche alarm forecasting task but even in simple alarm predictions.

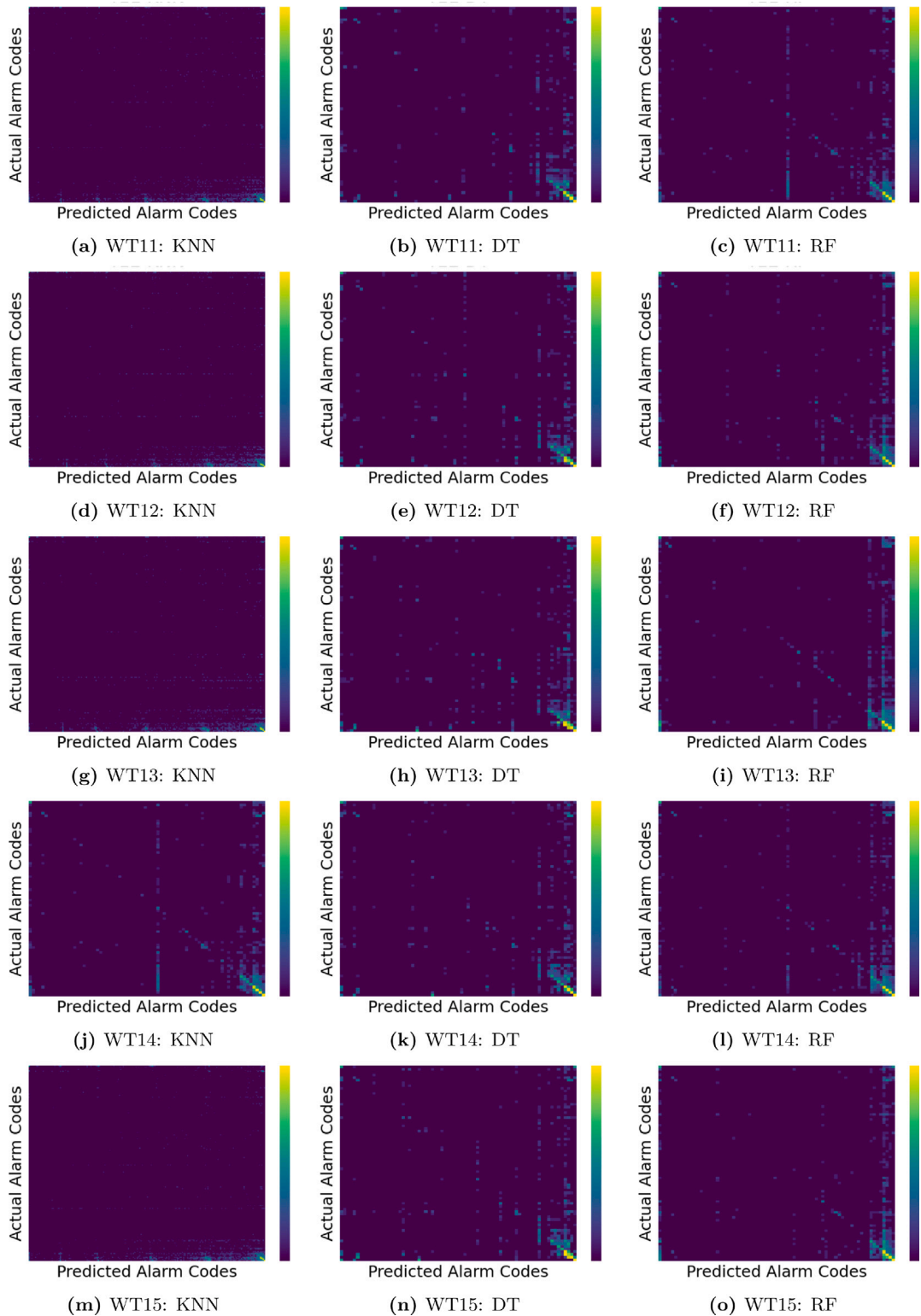

Fig. 25. FPAF (True positive—False negatives): FW 20 minutes.

3. With the achievement of forecasting incoming alarms, the AFC methodology makes way for the optimization and the implementation of the control system, for example, Predictive Maintenance (PdM)-based automated response, leading to the prevention of failures and thus boosting the sustainability of wind power.

4. Additionally, this study employed custom-designed novel preprocessing techniques to refine the SCADA data into a usable format. The preprocessing steps outlined in this research offer valuable guidance for future researchers seeking to improve WT dataset quality. By addressing the challenges posed by poor-quality data, these methods have the potential to unlock new avenues in WT research, as a significant portion of existing datasets remains underutilized due to their suboptimal quality.

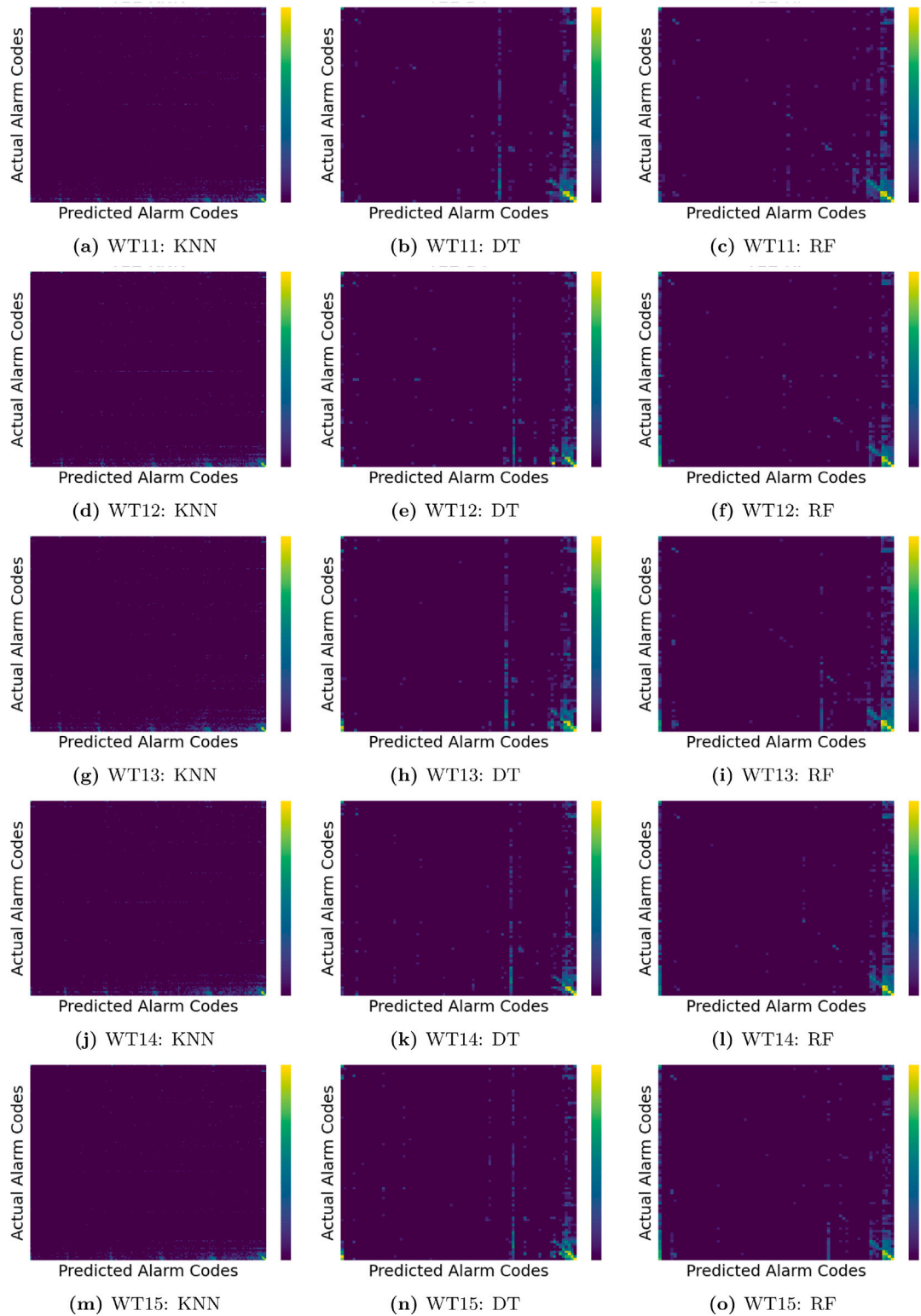

**Fig. 26.** FPAF (True positive—False negatives): FW 30 minutes.

**Note:** In the case of a 30 min forecast, 3 out of 5 results still remained at 40 to 50%, but since the least accurate case resulted in 30.6% accuracy, further increases in the FW size seemed impractical. Nevertheless, we are confident to present the results achieved for 30 min FW as one of the achievements, particularly because the datasets used for the AFC methodology were suboptimal in coherence with real-world conditions. Making use of relatively higher-quality datasets would definitely provide a comparative boost in accuracy.

## 6. Future directions

1. Exploration of Ensemble Methods: Combining multiple models to enhance predictive accuracy.

2. Data Augmentation: Incorporating additional data sources or employing data augmentation techniques to enrich the dataset.
3. Advanced Feature Engineering: Investigating techniques to extract more relevant and insightful information from the data.
4. SCADA Refinement: Using new SCADA systems with better frequency refinement to boost the accuracy of the AFC method.

**CRediT authorship contribution statement**

**Syed Shazaib Shah:** Writing – review & editing, Writing – original draft, Visualization, Validation, Software, Resources, Methodology, Investigation, Formal analysis, Data curation, Conceptualization. **Daoliang Tan:** Writing – review & editing, Supervision, Software, Resources, Project administration, Funding acquisition.

**Declaration of competing interest**

The authors declare the following financial interests/personal relationships which may be considered as potential competing interests: Syed Shazaib Shah reports financial support was provided by Beihang University. If there are other authors, they declare that they have no known competing financial interests or personal relationships that could have appeared to influence the work reported in this paper.

**Acknowledgments**

This work was supported by the China Scholarship Council (CSC) [No. 2021SLJ008269], which provided tuition and living expenses during the author's stay at Beihang University. The CSC scholarship did not include any research project or publication funding. The research period was supported School of Energy and Power Engineering, Beihang. Moreover, I am incredibly grateful to Goldwind for providing invaluable resources and insights.

**Appendix**

Figs. 18, 19 and 20 show the NaN value distribution post-NaN-removal operation, representing the analogous refinement result of SCADA. Due to the sake of clarity, the parameter count is restricted to the first 15. The figures provide a homogeneous spread of NaN logs across various parameters and a strong inter-turbine coherence, suggesting a strong relation between different parameters and dependency of multiple parameters on the same sensors. The inter-turbine coherence phenomena could also be related to ambient conditions, for instance, force majeure scenarios like extreme coherent gust, etc.

The alarm dataset consisted of 223 unique alarms; the prediction capabilities subjected to each alarm are different. Figs. 21, 22 and 23 show the AFC respective prediction performance for 10, 20 and 30 min forecast. Analytical view from these figures suggests AFC accuracy have an inherent direct proportional coherence to the given alarm frequency, where frequently recurring alarms exhibit higher prediction rates, while certain alarms were entirely overlooked by the model.

Fig. 24, 25 and 26 represent predicted versus actual alarm heat map correlation. The figures give valuable insights into the identification of alarms with their corresponding weightage. The prediction behavior can be leveraged to better understand model performance; referencing these figures, it is apparent DT and RF have a much tight prediction window, especially in case of 10 min forecast (Fig. 24). On that note, as major alarm frequency lies towards the end of denomination, the prediction spread tends to be much narrower (see bottom left corner for Fig. 24), however, the spread becomes more vertically even as the FW is increased from 10 min to 30, see Figs. 25 and 26, showcasing the reasoning for the degrading effect exhibited in Fig. 13.

**Data availability**

Data will be made available on request.